\documentclass[manuscript,acmsmall,screen,nonacml]{acmart}
\usepackage{multirow}
\usepackage{makecell,booktabs,threeparttablex,cleveref,graphicx}
\usepackage{makecell}
\usepackage{subcaption}
\usepackage{longtable}
\usepackage{array}

\newcolumntype{C}[1]{>{\centering\arraybackslash}p{#1}} 
\newcolumntype{L}[1]{>{\raggedright\arraybackslash}p{#1}} 
\newcolumntype{R}[1]{>{\raggedleft\arraybackslash}p{#1}} 
\AtBeginDocument{%
  }
\crefname{table}{Tab.}{Tab.}

\setcopyright{acmlicensed}
\copyrightyear{2026}
\acmYear{2026}
\acmDOI{XXXXXXX.XXXXXXX}
\acmConference[Conference acronym 'XX]{Make sure to enter the correct
  conference title from your rights confirmation email}{June 0X--0X,
  20XX}{Woodstock, NY}
\acmISBN{978-1-4503-XXXX-X/2018/06}

\begin{document}

\title{Deep Neural Networks Inspired by Differential Equations}
\author{Yongshuai Liu}
\email{lys_bnu@mail.bnu.edu.cn}
\author{Lianfang Wang}
\email{202431130071@mail.bnu.edu.cn}
\author{Kuilin Qin}
\email{202321130113@mail.bnu.edu.cn}
\author{Qinghua Zhang}
\email{qhzhang@mail.bnu.edu.cn}
\author{Faqiang Wang}
\email{fqwang@bnu.edu.cn}
\author{Li Cui}
\email{licui@bnu.edu.cn}
\author{Jun Liu}
\email{jliu@bnu.edu.cn}
\author{Yuping Duan}
\authornote{Corresponding author.}
\email{doveduan@gmail.com}
\affiliation{
  \institution{School of Mathematical Sciences, Beijing Normal University}
  \city{Beijing}
  \country{China}}

\author{Tieyong Zeng}
\affiliation{
  \institution{Department of Mathematics, The Chinese University of Hong Kong}
  \city{Shatin}
  \country{Hong Kong}}
\email{zeng@math.cuhk.edu.hk}

\renewcommand{\shortauthors}{Liu et al.}

\begin{abstract}
Deep learning has emerged as an important technology in fields such as computer vision, scientific computing, and dynamical systems, thereby driving major advancements across these domains.  However, neural networks persistently face challenges related to theoretical understanding, interpretability, and generalization. Thus, researchers are increasingly adopting a differential equations perspective to propose a unified theoretical framework and systematic design methodologies. We establish a comprehensive taxonomy of the literature from two distinct perspectives. First, we classify the models based on their mathematical formulations, covering ordinary, partial, and stochastic differential equations (ODEs, PDEs, and SDEs). Second, we categorize them by their methodological roles, distinguishing between equation-guided networks and networks designed for solving equations. Moreover, we conduct a qualitative meta-analysis supported by consolidated evidence summaries from existing literature to illustrate their characteristics and performance. Finally, we explore promising research directions in integrating differential equations with deep learning to offer new insights for developing computational methods with improved interpretability and generalization.
\end{abstract}

\ccsdesc[500]{Computing methodologies~Philosophical/theoretical foundations of artificial intelligence}
\ccsdesc[500]{Networks~Network design principles}
\ccsdesc[500]{Mathematics of computing~Differential equations}

\keywords{Deep neural network, ordinary differential equation, stochastic differential equation, network structure, dynamic process}


\maketitle

\section{Introduction}
Deep learning has emerged as a transformative paradigm in modern artificial intelligence, driven by its success in tasks such as computer vision \cite{o2019deep,chai2021deep,voulodimos2018deep}, natural language processing \cite{otter2020survey,young2018recent,collobert2008unified}, autonomous driving \cite{muhammad2020deep,badjie2024adversarial} and medical measurement \cite{aggarwal2021diagnostic,singh20203d}. These achievements are largely attributed to the confluence of three key factors: the proliferation of large-scale annotated datasets \cite{veit2017learning}, the rapid advancement in computational hardware \cite{chen2020deep,sze2017hardware} (e.g., GPUs and TPUs), and the continuous evolution of neural network architectures~\cite{liu2017survey,unal2022evolutionary}. Despite these empirical successes, deep learning still faces challenges in interpretability and generalization~\cite{belkin2019reconciling,10.1145/3594869}, which limit its applicability in domains requiring rigorous safety and reliability guarantees~\cite{doshi2017towards,rudin2019stop}. 

ODEs and SDEs have emerged as theoretical frameworks providing novel insights and mathematical tools for tackling the challenges of interpretability and generalization in deep learning. Pioneering work by E et al.~\cite{weinan2017proposal} conceptualized deep neural networks as discrete dynamic systems, framing the training phase as an optimal control problem. This perspective catalyzed the development of network architectures grounded in deterministic differential equations, encompassing ODEs and partial differential equations (PDEs), and leading to substantial advancements~\cite{chen2018neural,lu2018beyond,ruthotto2020deep,gomez2017reversible}. Stability in ODE-based models is often achieved by enforcing Lipschitz or contraction conditions on the function $f$, which ensures reliable forward and backward propagation and mitigates gradient explosion or vanishing. Overall, existing research can be broadly categorized into two main directions: 1) leveraging differential equation frameworks, including ODEs, PDEs, and SDEs, for the theoretical interpretation of deep networks~\cite{weinan2017proposal,zhang2017beyond,xie2017aggregated,larsson2016fractalnet,targ2016resnet,huang2017densely,gomez2017reversible,behrmann2019invertible,alt2023connections}; 2) guiding the design of network architecture with DEs~\cite{lu2018beyond,ruthotto2020deep,zhang2020forward,he2019ode,luo2022rethinking,kong2022hno,greydanus2019hamiltonian,ephrath2020leanconvnets,kag2022condensing,wang2025convection}. 

The intrinsic interpretability of ODEs arises from their deterministic nature; given an initial state, systems evolve along a single, predictable trajectory. This transition from discrete layer stacking to continuous evolution enhances the model's physical intuitiveness and interpretability. For example, ODE-based models like ResNets~\cite{he2016deep} and LM-ResNets~\cite{lu2018beyond} interpret inter-layer evolution as a discretization of continuous flows, whereas PDE-based models such as PDE-CNNs~\cite{ruthotto2020deep} encode geometric structures and local features through spatial derivatives. Additionally, deterministic DEs facilitate deterministic generative modeling, including Neural ODEs \cite{chen2018neural} and flow matching methods \cite{lipman2022flow,lipman2024flow}. These frameworks generate data by solving initial value problems or construct continuous trajectories between distributions by deriving velocity fields through optimal transport. Furthermore, this continuous perspective extends to scientific machine learning through Neural PDE solvers, which operate under specified equation constraints to find numerical solutions for complex physical systems. These approaches illustrate the role of deterministic DEs in the unification of neural architectures, generative modeling and equation solving.

SDEs extend ODEs by incorporating a stochastic noise term, transforming deterministic systems into stochastic ones. This stochasticity induces trajectory variability from identical initial states, capturing richer distributional variations to improve generalization and enable the generation of diverse, high-quality samples. Randomness is pervasive in deep neural networks, arising from weight initialization~\cite{narkhede2022review,kumar2017weight}, minibatch sampling~\cite{csiba2018importance}, and stochastic regularization methods such as Dropout~\cite{baldi2013understanding,salehin2023review,labach2019survey,gal2016theoretically} and Gaussian noise injection~\cite{he2019parametric,camuto2020explicit,camuto2021understanding}. These strategies influence optimization dynamics, robustness, and generalization. Recent studies employ SDEs to rigorously analyze this randomness, forming two primary research trajectories. One trajectory focuses on stochastic regularization: explicitly embedding randomness via SDE frameworks (e.g., Dropout~\cite{hinton2012improving,srivastava2014dropout} and stochastic depth~\cite{huang2016deep}) to provide coherent mathematical models for network evolution under perturbations~\cite{lu2018beyond,galeati2022regularization,wang2025convection}. Another trajectory employs SDEs to model network dynamics~\cite{song2020score,batzolis2021conditional,luo2023image,cao2025generative,cao2023exploring,liu2019neural,jia2019neural,li2020scalable,kidger2021neural,oh2024stable}, comprising two categories: Neural stochastic differential equations (Neural SDEs)~\cite{liu2019neural} and SDE-guided generative models, exemplified by Denoising Diffusion Probabilistic Models (DDPMs)~\cite{song2020score}.

\begin{figure*}[htbp]
    \centering
    \includegraphics[width=0.7\linewidth]{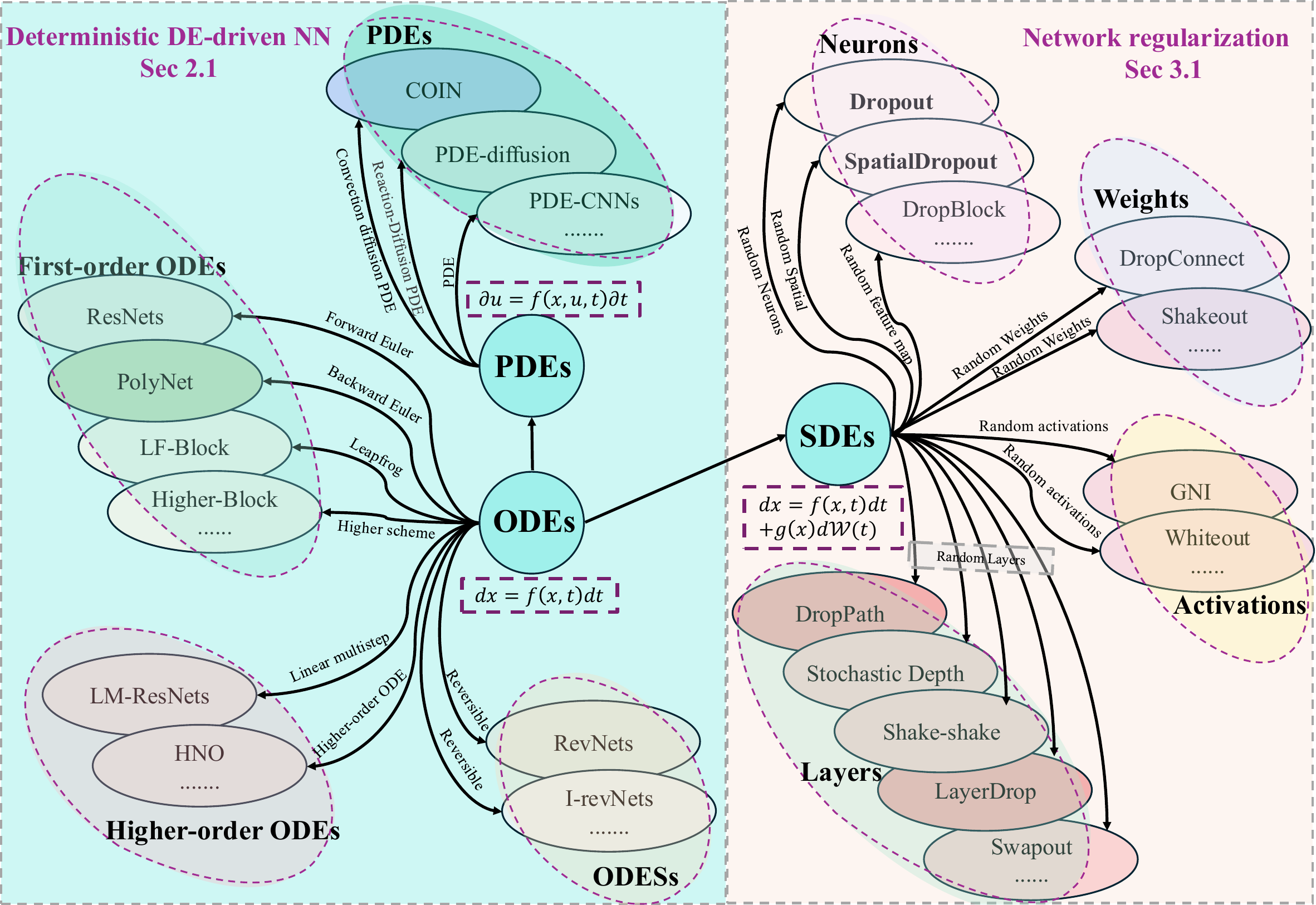}
    \caption{Overview of deterministic DE-driven neural network architectures (left, Sec.\ref{sec:DDEs_NN}) and stochastic regularization methods inspired by SDEs (right, Sec.\ref{Sec:Network_regularition}).}
    \label{fig:nns}
\end{figure*}

\begin{figure*}[htbp]
    \centering
    \includegraphics[width=0.7\linewidth]{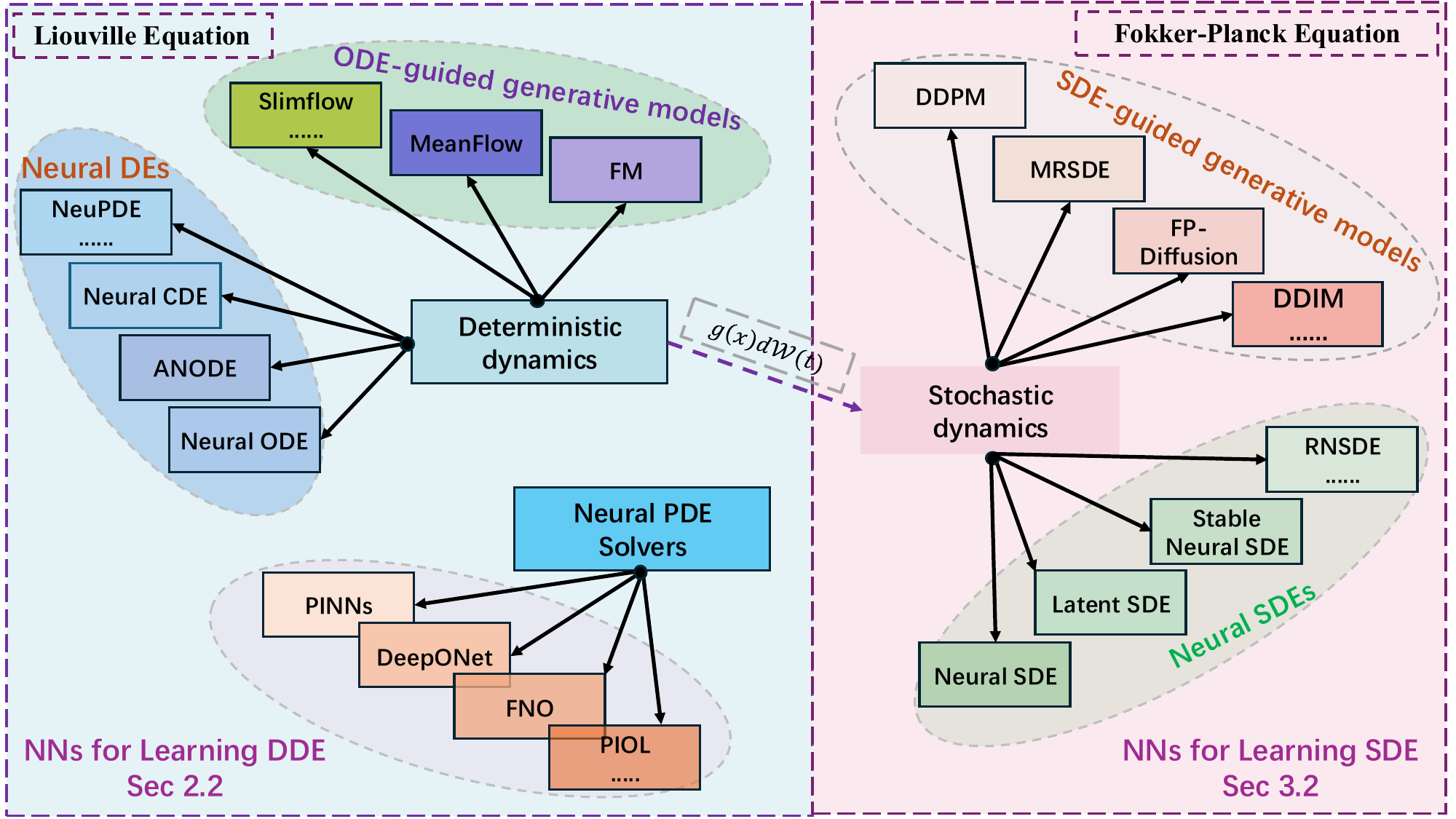}
    \caption{Neural Networks for Learning Differential Equations: Deterministic and Stochastic. Specifically, the left panel illustrates networks tasked with learning deterministic dynamical simulations and numerical solutions of PDEs (Sec.\ref{Sec:modeling_DD}), whereas the right panel focuses on networks learning stochastic dynamics (Sec.\ref{sec:dynamic_SDEs}).}
    \label{fig:dl}
\end{figure*}

Taking a broader perspective, ODEs, PDEs, and SDEs collectively span a systematic theoretical hierarchy, ranging from deterministic temporal evolution to complex stochastic spatiotemporal systems. As illustrated in Fig.~\ref{fig:nns}, these differential equations serve as rigorous foundations for network design and regularization. Furthermore, Fig.~\ref{fig:dl} demonstrates how deep learning leverages these continuous dynamics to either drive data evolution for generative modeling or compute accurate numerical solutions. We defer a comprehensive review of the fundamental neural architectures and mathematical prerequisites to Section S1 of the online supplementary material, thereby ensuring the main narrative remains concise and focused.

This survey limits its scope to model-level paradigms whose architectural topologies, generative continuous dynamics, or algorithmic equation-solving procedures are explicitly governed by ODEs, PDEs, or SDEs. Consequently, layer-level innovations such as Structured State-Space Models (e.g., S4~\cite{gu2022efficiently}, Mamba~\cite{gu2023mamba}) and Liquid Time-Constant Networks (LTCs)~\cite{Hasani2021} fall outside our primary focus. We refer interested readers to \cite{zhao2026understanding} for a comprehensive review of these layer-centric architectures. Existing surveys predominantly categorize models by specific domains, techniques, or isolated equation types, as detailed in \cref{tab:survey_comparison}. For the systematic organization of this field, we introduce a comprehensive framework based on dual-perspective classification criteria. First, we classify the literature by mathematical foundations: ODEs, PDEs, and SDEs. Second, we categorize models by methodological roles into two main areas. The first regime, Networks Guided by Equations, encompasses DE-driven architectures and stochastic regularization. The second regime, Networks for Learning Equations, subdivides into neural solvers for finding numerical solutions and networks learning data-driven dynamics, such as Neural DEs and DE-guided generative models. 

In summary, this survey provides a systematic taxonomy and a holistic review of deep learning paradigms underpinned by differential equations. Specifically, our methodological classification successfully organizes the literature into two primary functional regimes. The first regime, Networks Guided by Equations, encompasses differential equation-driven architectures and stochastic regularization techniques. The second regime, Networks for Learning Equations, is further subdivided into neural numerical solvers and data-driven continuous dynamics. Beyond this structural taxonomy, we perform a rigorous qualitative meta-analysis of these DE-centric models across various standardized evaluation benchmarks. Finally, we synthesize prevailing open challenges and chart prospective research trajectories to navigate this rapidly evolving domain.

\begin{table*}[htbp]
\centering
\renewcommand{\arraystretch}{1.0} 
\caption{Comparison with Existing Surveys: Scope and Coverage.}
\label{tab:survey_comparison}
\resizebox{\textwidth}{!}{
\begin{tabular}{l l l }
\toprule
\textbf{Year\&Reference} & \textbf{Scope of Models}  & \textbf{Summary}  \\
\midrule
2022~\cite{santos2022avoiding} &Network regularization & \makecell[l]{This survey paper explores recent regularization techniques to prevent overfitting in CNNs,\\ categorizing them into data augmentation, internal structure changes, and label transformations.}\\
\midrule
2024~\cite{yang2023diffusion} &Diffusion models &\makecell[l]{This survey provides a comprehensive taxonomy of diffusion models, categorizing advancements\\ into efficient sampling, improved likelihood estimation, and data generalization.}\\
\midrule
2024~\cite{cao2024survey} &Diffusion models &\makecell[l]{This survey systematically elucidates the developmental trajectory of generative diffusion models,\\ categorizing advancements in theoretical formulations and algorithmic efficiency.}\\
\midrule
2025~\cite{huang2025diffusion} & \makecell[l]{Diffusion models\\  (Image Editing)}&\makecell[l]{This survey systematically categorizes and reviews text-to-image diffusion model-based image editing\\ techniques based on their learning requirements—training-free, fine-tuning, and feed-forward methods.}\\
\midrule
2025~\cite{cao2025controllable} &\makecell[l]{Diffusion models\\ (text2image)}& \makecell[l]{This survey systematically categorizes techniques for controlling text-to-image diffusion models, focusing\\ on prompt engineering, spatial guidance, and personalized concept injection.}\\
\midrule
2024~\cite{lipman2024flow} &Flow matching & \makecell[l]{This tutorial bridges the mathematical theory of flow Matching and variant models with practical\\ PyTorch implementations to demonstrate efficient generative modeling.}\\
\midrule
2025~\cite{li2025flow} &\makecell[l]{Flow matching \& biology \\and life sciences} &\makecell[l]{The paper outlines how flow matching offers efficient generative solutions for various computational\\ biology challenges ranging from drug discovery to clinical diagnostics.}\\
\midrule
2025~\cite{huang2025partial} & \makecell[l]{DNNs \\for solving PDEs} &\makecell[l]{This survey reviews deep learning methods for solving PDEs categorizing them into models with equation \\constraints such as PINNs and operator learning networks.}\\
\midrule
2022~\cite{kidger2022neural} & Neural differential equations&\makecell[l]{This thesis provides an in depth survey of neural deterministic differential equations.}\\
\midrule
2025~\cite{oh2025comprehensive} &Neural differential equations & \makecell[l]{This survey provides a comprehensive review of neural differential equation based methods for time series\\ analysis.}\\
\midrule
\textbf{Ours} &\makecell[l]{DE-DNN\\ Diffusion models\\ Flow matching \\ Neural differential equations \\ DNN for solving PDEs} & \makecell[l]{This survey establishes a unified framework for deep learning models using DEs and explores the\\ bidirectional synergy between DE-guided network design and networks for learning equations.}\\
\bottomrule
\end{tabular}
}
\end{table*}

\section{From Deterministic DEs to Deep Neural Networks}
Since the pioneering work by E et al. \cite{weinan2017proposal} that reinterpreted the essence of neural networks from a dynamical systems perspective, DE-guided neural networks design has emerged as  an active research direction \cite{sun2018stochastic,zhang2017polynet,larsson2016fractalnet,ruthotto2020deep,haber2017stable,chen2018neural,gomez2017reversible,lu2018beyond}. This perspective not only established a rigorous mathematical framework for deep learning but also led to new architectures that combine robustness and interpretability, substantially enhancing model adaptability in complex real-world applications.

To systematically incorporate the mathematical rigor of deterministic DEs into deep learning, current mainstream approaches can be categorized into two key paradigms:
\begin{itemize}
    \item \textbf{Deterministic DE-Driven Architectures (Networks Guided by Equations):} Exemplified by ResNets~\cite{he2016deep} and LM-ResNets~\cite{lu2018beyond}, this paradigm formulates neural network forward propagation as solving initial value problems. By treating layers as discrete deterministic DE approximations via Euler or Runge-Kutta schemes~\cite{kidger2020neural}, this perspective elucidates the networks' function approximation capabilities and guides novel architectural design.
    \item \textbf{Networks for Learning Deterministic Equations:} This paradigm deploys deep learning to directly learn or compute continuous dynamical systems. For data representation, deterministic dynamics models like Neural ODEs~\cite{chen2018neural} and ODE-guided generative models~\cite{lipman2022flow} describe the continuous evolution of data in latent space, leveraging temporal continuity for flexible trajectory interpolation and precise gradient computation. Extending beyond data modeling, this paradigm also encompasses Neural PDE solvers~\cite{lu2021learning,li2020fourier}, which constrain networks with specific physical laws to provide efficient numerical solutions for complex physical systems.
\end{itemize}

\subsection{Deterministic DE-Driven Network Architectures}\label{sec:DDEs_NN}
We first present a structured classification of deterministic DE-driven approaches for deep learning architecture design. A detailed summary of representative works is provided in \cref{tab:DEs_DL}, and the corresponding architectural illustrations are presented in Fig. S1 of the online supplementary material.

\begin{table*}[htbp]
\caption{Representative deep learning architectures driven by deterministic DEs, categorized by first-order ODEs, higher-order ODEs, DESs, and PDEs.}
\label{tab:DEs_DL}
\centering
\resizebox{0.8\textwidth}{!}{
\begin{tabular}{llll}
\toprule
\textbf{Category} & \textbf{Model} & \textbf{Formulation}&\textbf{Year} \\
\midrule
\multirow{11}{*}{\textbf{First-order ODEs}} & ResNets~\cite{he2016deep} &   $x_{n+1}=x_n+f(x_n)$ &2016\\
 \cmidrule{2-4}
 & RiR~\cite{targ2016resnet} & $x_{n+1}=x_n+f(x_n,t_n),\,t_{n+1}=g(x_n,t_n)$&2016   \\
  \cmidrule{2-4}
& FractalNet~\cite{larsson2016fractalnet} & $x_{n+1}=k_1x_n+k_2(k_3x_n+f_1(x_n))+f_2(k_3x_n+f_1(x_n))$ &2016 \\
  \cmidrule{2-4}
 & ResNeXt~\cite{xie2017aggregated} & $x_{n+1}=x_n+\sum\limits_i f_i(x_n)$&2017 \\
  \cmidrule{2-4}
 & PolyNet~\cite{zhang2017polynet} &   $x_{n+1} = (I-\Delta t f)^{-1} x_n$ &2017 \\
  \cmidrule{2-4}
 & LF-Block~\cite{he2019ode}& $x_{n+1}=x_{n-1}+2h f(x_n)$&2019 \\
 \cmidrule{2-4}
 & Higher-block~\cite{luo2022rethinking} & 
 \makecell[l]{
 $k_1=f(x_n),\,k_2=f(x_n+\frac12k_1),$\\
 $k_3=f(\frac12k_2+x_n),\,k_4=f(k_3+x_n)$\\
 $x_{n+1}=x_n+k_2\,\text{(Midpoint)}$\\
 $x_{n+1}=x_n+\frac{h}{2}(k_1+f(x_n+k_1))$\,\text{(RK2)}\\
$x_{n+1}=x_n+\frac16(k_1+k_2+k_3+k_4)\,\text{(RK4)}$\\
 }&2022\\
\midrule
\multirow{2}{*}{\textbf{Higher-order ODEs}} & LM-ResNets~\cite{lu2018beyond}  &  $x_{n+1}=(1-k)x_n+kx_{n-1}+f(x_n)$ &2018\\
 \cmidrule{2-4}
 & HNO~\cite{kong2022hno}  &  $x_{n+1}=(\xi_n)_0 x_n+(\xi_n)_1 x_{n-1}+\cdots$ &2022\\
\midrule
\multirow{3}{*}{\textbf{DE Systems}} & RevNets~\cite{gomez2017reversible} &  $y_{n+1}=y_n+f(x_n),\, x_{n+1}=x_n+g(y_{n+1})$&2017\\
  \cmidrule{2-4}
 & HamiltonianNet~\cite{chang2018reversible}&$y_{n+1}=y_n+f(x_n),\,x_{n+1}=x_n-g(y_{n+1})$&2018\\
 \cmidrule{2-4}
 & i-RevNet~\cite{behrmann2019invertible}& $y_{n+1}=y_n+f(x_n),\,x_{n+1}=P(x_n)+g(y_{n+1})$&2019\\
\midrule
\multirow{9}{*}{\textbf{PDEs}} &PDE-CNNs~\cite{ruthotto2020deep}&
\makecell[l]{
$F_{\text{sym}}(\theta,x)=-K(\theta)^T\sigma(\mathcal{N}(K(\theta)x,\theta))$\\
$x_{n+1}= x_n + F_{\text{sym}}(\theta_n,x_n)\,\text{(Parabolic CNN)}$\\
$x_{n+1}=2x_n-x_{n-1}+F_{\text{sym}}(\theta_n,x_n)\,\text{(Second-order CNNs)}$\\
}&2020\\
\cmidrule{2-4}
&LeanConvNets~\cite{ephrath2020leanconvnets}&$x_{n+1}=x_n+f(x_n,\theta_n)$&2020\\
 \cmidrule{2-4}
 &PDE-diffusion~\cite{kag2022condensing}&$\frac{\partial H}{\partial t} = \nabla\cdot(D\nabla H)+\nabla\cdot(vH)+f(I)$&2022\\
 \cmidrule{2-4}
 & PDEs-DNNs~\cite{alt2023connections}&$x_{n+1}=x_n-K^T\phi(Kx_n)$&2023\\
 \cmidrule{2-4}
 &COIN~\cite{wang2025convection} &\makecell[l]{
$\frac{\partial u(x,t)}{\partial t}=v(x,t)\cdot\nabla u(x,t),\,t\in[0,T-1]$\\
$\frac{\partial u(x,t)}{\partial t} = \sigma^2\Delta u(x,t),\,t\in[T-1,T]$\\
 $u(x,0)=f(x)$\\
 }&2025\\
\bottomrule
\end{tabular}
}
\end{table*}

\subsubsection{\textbf{ResNets}~\cite{he2016deep}} ResNets revolutionized deep learning by introducing skip connections to overcome the degradation problem in very deep DNNs. The fundamental building block implements a residual transformation through the update rule:
\begin{equation}\label{ResNet1}
    x_{n+1} = x_n + f(x_n),
\end{equation}
where $x_n$ is the layer $n$ feature representation and $f$ is a learnable convolutional transformation. Such skip connections stabilize gradient propagation and facilitate the training of ultra-deep architectures like ResNet-50.

Beyond their empirical success, ResNets have been rigorously analyzed from a dynamical systems perspective. Haber et al. \cite{haber2017stable} and E et al. \cite{weinan2017proposal} established that the residual update in \eqref{ResNet1} can be viewed as a forward Euler discretization of an underlying continuous-time ODE:
\begin{equation}\label{ODERN}
    \frac{dx(t)}{dt} = v(x,t),\,\, x(0)=x_0,\,\, t\in [0,T],
\end{equation}
where $v(x,t)$ denotes a velocity field satisfying $v(x, t) = f(x) / \Delta t$. This connection enables the numerical analysis of network stability and inspires continuous-depth models using ODE solvers for adaptive computation, motivating ResNet variants and advancing our theoretical understanding.

Since the introduction of ResNets, several variants have been proposed to enhance its performance. Architecturally, Wide ResNet (WRN)~\cite{zagoruyko2016wide} and ResNeXt~\cite{xie2017aggregated} increase network width and cardinality, while ResNet-S and ResNet-D~\cite{zhang2020forward} focus on optimizing downsampling paths. These adaptations balance depth and width for more efficient training. Numerically, variants like Residual-in-Residual Network (RiR)~\cite{targ2016resnet}, Polynomial Network (PolyNet)~\cite{zhang2017polynet}, and Fractal Network (FractalNet)~\cite{larsson2016fractalnet} explore different formulations to improve stability and expressiveness. Methods such as leapfrog block (LF-Block)~\cite{he2019ode},  Runge-Kutta-2 block (RK2-Block)~\cite{he2019ode}, and higher-order RK4/RK8~\cite{luo2022rethinking} integrate time-stepping techniques from numerical analysis to enhance learning accuracy and convergence.

\subsubsection{\textbf{Linear Multistep ResNets}~\cite{lu2018beyond}} LM-ResNets introduce multistep dependencies inspired by Linear Multistep Methods (LMMs). Unlike standard ResNets that update solely from the current state $x_n$, LM-ResNets compute $x_{n+1}$ as a neural-parameterized nonlinear function of $x_n$ and past states (e.g., $x_{n-1}$). The layer update follows the general form of LMMs:
\begin{equation*}
    x_{n+1} = (1-k_n)x_n+k_n x_{n-1}+f_n(x_n),
\end{equation*}
where $k_n\in\mathbb{R}$ is a trainable parameter. LM-ResNets can be interpreted as a high-order ODE and written as:
\begin{equation*}
    f(x_n,t)=(1+k_n)\dot{x}_n + (1-k_n)\frac{\Delta t}{2}\ddot{x}_n.
\end{equation*}
Subsequently, HNO~\cite{kong2022hno} generalizes deep unfolding networks as ODE solvers, employing linear multistep methods to accelerate convergence and improve accuracy. Concurrently, DenseNet~\cite{huang2017densely} introduces a densely connected topology that concatenates features from all preceding layers, enhancing both feature reuse and gradient flow.

\subsubsection{\textbf{Reversible Residual Networks (RevNets)} \cite{gomez2017reversible}} Reversible architectures, such as RevNets \cite{gomez2017reversible}, invertible Residual Networks (i-ResNets)~\cite{behrmann2019invertible}, and invertible Reversible Networks (i-RevNets)~\cite{jacobsen2018revnet}, address the memory bottleneck of traditional ResNets by reconstructing activations during backpropagation. This eliminates the need to store intermediate states within residual blocks, which can be formally expressed as:
\begin{equation*}\label{Eq:RevNet}
y_{n+1} = y_n + f(x_n),\,
x_{n+1} = x_n + g(y_{n+1}),
\end{equation*}
where $f$ and $g$ are learnable residual functions. The inputs are reconstructed as:
\begin{equation*}
 x_n = x_{n+1} - g(y_{n+1}),\,
y_n = y_{n+1} - f(x_n).
\end{equation*}
RevNets reduce memory usage with limited performance loss but incur higher computational cost. I-ResNets~\cite{behrmann2019invertible} achieve full-network invertibility for both generative and discriminative tasks, while i-RevNets~\cite{jacobsen2018revnet} add nonlinear mappings to further improve memory efficiency. Hamiltonian Networks~\cite{greydanus2019hamiltonian,chang2018reversible} extend RevNets by modeling reversible dynamical systems, enabling stable, memory-efficient training and accurate physical simulations.

\subsubsection{\textbf{PDE-CNNs}~\cite{ruthotto2020deep}} In \eqref{ResNet1}, the function $f$ is treated as a black box without structural constraints. For tasks involving structural data (e.g., speech, image, or video processing) where local spatial interactions are crucial, layers are standardly instantiated via convolutions. Ruthotto et al.~\cite{ruthotto2020deep} proposed a PDE-based framework that interprets the forward propagation of residual CNNs as a continuous-time nonlinear diffusion process. Formally, given the initial state as the input, the underlying continuous dynamics are modeled via the forward Euler method:
\begin{equation*}
    \frac{\partial x(\theta,t)}{\partial t} = f(\theta,x,t),\, x(\theta,0)=x_0,\,f(\theta,x,t) = K_2(\theta^{(3)})\,\sigma\Big( \mathcal{N}\big(K_1(\theta^{(1)})Y,\theta^{(2)} \big)
 \Big)
\end{equation*}
where $f(\theta,x,t)$ represents convolution, nonlinearity, and normalization, with specific parameterizations for each operation. Specifically, $K_1(\theta^{(1)})$ and $K_2(\theta^{(3)})$ denote convolutional operators parameterized by $\theta^{(1)}$ and $\theta^{(3)}$, respectively. $\mathcal{N}$ represents a normalization layer parameterized by $\theta^{(2)}$, and $\sigma$ denotes a nonlinear activation function (e.g., ReLU). PDE-CNNs~\cite{ruthotto2020deep} encompass parabolic and hyperbolic models, including Hamiltonian and second-order variants. While PDE formulations improve stability and interpretability, deep architectures often incur high computational costs. To mitigate this, LeanConvNets~\cite{ephrath2020leanconvnets} and Global Feature Layers~\cite{kag2022condensing} reduce structural complexity. Furthermore, extending PDE frameworks to ResNets and RNNs enhances parameter efficiency~\cite{alt2023connections}, and Convection-Diffusion Networks (COIN)~\cite{wang2025convection} further demonstrate robust generalization against perturbations.

Deterministic DE-driven neural architectures can be understood through a unified trade-off among interpretability, stability, expressivity, and computational cost, governed by how the underlying dynamical system constrains state evolution. 
First-order ODE-inspired models use simple single-step updates, offering strong stability and efficiency but limited expressivity, requiring deep compositions for complex mappings. 
Higher-order ODEs introduce multi-step dependencies, increasing degrees of freedom and representation capacity at the cost of additional state and computation. 
DE systems impose structured dynamics (e.g., invertibility or conservation laws), improving interpretability and memory efficiency while potentially limiting flexibility. 
PDE-based architectures extend dynamics to the spatial domain, enabling effective spatio-temporal modeling but incurring higher computational cost due to spatio-temporal discretization. 
Overall, improving expressivity and interpretability generally requires relaxing structural simplicity and increasing computational complexity, highlighting the need to match the equation type to the target task.

\begin{table*}[htbp]
\caption{A Taxonomy of Representative Neural Differential Equation Models: Deterministic vs. Stochastic Formulations. Deterministic formulations emphasize computational efficiency and memory savings but are constrained by limited flexibility, while stochastic models enhance expressivity through noise injection at the cost of increased computational overhead.}
\label{tab:De_dN}
\centering
\resizebox{\textwidth}{!}{
\begin{tabular}{lllll}
\toprule
\textbf{Model} & \textbf{Formulation} &\textbf{Driving Mechanism}& \textbf{Typical Solvers} &\textbf{Year}\\
\midrule
\multicolumn{5}{c}{\textbf{(a) Neural Deterministic Differential Equations}} \\ 
\midrule
 Neural ODEs~\cite{chen2018neural} & $dx(t)=f_\theta(x,t)dt$ & Vector Field & Implicit Adams &2018\\
\midrule
  ANODEs~\cite{dupont2019augmented} & 
 \makecell[l]{
$\dfrac{d}{d t}\begin{pmatrix}
         x(t)\\a(t)
     \end{pmatrix}=f_\theta \left(\begin{pmatrix}
         x(t)\\a(t)
     \end{pmatrix},t \right) $
}&Vector Field &Dopri5 &2019 \\
\midrule
SONODEs~\cite{norcliffe2020second} & $\ddot{x}(t)=f_\theta(x,\dot{x},t)$& Second-order Field&  Dopri5 &2020\\
\midrule
Neural jump ODE~\cite{herrera2020neural}&
 \makecell[l]{
$\tilde{h}_i = \text{ODESolve}(f_\theta, h_{i-1}, (t_{i-1}, t_i))$\\
$h_i = \text{RNNCell}(\tilde{h}_i, x_i)$
 }&Vector Field + Jumps &Euler &2021\\
 \midrule
NeuPDE~\cite{sun2020neupde} & $\frac{\partial u(t, x)}{\partial t} = \mathcal{N}_\theta (\mathcal{D}(u))$ & Vector Field + Jumps & Runge-Kutta &2020\\
\midrule
Neural CDEs~\cite{kidger2020neural} &$dz(t)=f_\theta(z)\,dX(t),\,X(t)\, \text{is control path}$&Control Path & Dopri5&2020\\
\midrule
Neural RDEs~\cite{morrill2021neural} &\makecell[l]{
$dz(t)=f_\theta(z(t))d\text{ LogSig}_{[r_i,r_{i+1}]}(X)$
}&\makecell[l]{Rough Path\\ (Log-signature)} & Dopri5 & 2021\\
\cmidrule(lr){2-4}
Neural FDEs~\cite{coelho2025neural}&$D^\alpha z(t) = f_\theta(z,t),\,0<\alpha\le 1$&Fractional Derivative & Predictor-Corrector&2025\\
\midrule
\multicolumn{5}{c}{\textbf{(b) Neural Stochastic Differential Equations}} \\ 
\midrule
Neural SDEs~\cite{liu2019neural} &$dz(t)=f_\theta (z,t)dt+g_\theta(z,t)d\mathcal{W}(t)$&Additive Noise & Euler-Maruyama&2019   \\
\midrule
Neural Jump SDEs~\cite{jia2019neural} & $dz(t) = f_\theta(z,t)dt+\omega_\theta(z,k,t)\cdot dN(t)$&Additive Noise & Jump Solver &2019\\
\midrule
Latent SDE~\cite{li2020scalable}&
\makecell[l]{
$d\tilde{z}(t)=h_\theta(\tilde{z},t)dt+\sigma(\tilde{z},t)d\mathcal{W}(t)\,(\text{prior})$\\
$dz(t)=h_\phi(z,t)dt+\sigma(z,t)d\mathcal{W}(t)\,(\text{approximate posterior})$
}&Multiplicative Noise &Milstein&2020\\
\midrule
Neural SDEs as GAN~\cite{kidger2021neural} &\makecell[l]{
$dx(t)=\mu_\theta(x,t)dt+\sigma_\theta(x,t)d\mathcal{W}(t)\,(\text{generator})$\\
$dh(t)=f_\phi(h,t)dt+g_\phi(h,t)dY(t)\,(\text{discriminator})$
}&Multiplicative Noise &Midpoint &2021 \\
\midrule
Neural Covariance SDE~\cite{li2022neural}& $d\phi_t(x) = -\frac12 \int \sum_t(x,x')\nabla\phi_t(x')dx'dt+\sqrt{\sum_t(x,x)}\,d\mathcal{W}(t)$&Covariance-shaped Noise &Euler–Maruyama &2022\\
\midrule
RNSDE~\cite{park2022riemannian} & $dx(t) = V(x,t)dt+\sum_{i=1}^m G_i(x,t)\,d\mathcal{W}(t)$ &Multiplicative Noise &Euler–Maruyama &2022\\
\midrule
Stable Neural SDEs~\cite{oh2024stable} & \makecell[l]{
Neural LSDE: $dz(t)=\gamma(\bar{z};\theta_\gamma)dt+\sigma(\theta_\sigma;t)d\mathcal{W}(t)$\\
Neural LNSDE: $dz(t)=\gamma(\bar{z},t;\theta_\gamma) \, dt+\sigma(\theta_\sigma;t)\,z\,d\mathcal{W}(t)$\\
Neural GSDE: $\frac{dz(t)}{z(t)}=\gamma(\bar{z},t;\theta_\gamma)dt+\sigma(\theta_\sigma;t)d\mathcal{W}(t)$\\
}&\makecell[l]{Additive Noise\\ Multiplicative Noise\\ Geometric Noise} &\makecell[l]{ Euler–Maruyama\\Euler–Maruyama\\Euler–Maruyama} &2024\\
\bottomrule
\end{tabular}
}
\end{table*}

\begin{table*}[!t]
\caption{Overview of ODE-guided and SDE-guided generative models: Deterministic speed vs. Stochastic diversity. ODEs enable rapid inference but limit sample diversity, whereas SDEs guarantee robust distribution coverage at the cost of slow sampling.}
\label{tab:flow_model}
\renewcommand{\arraystretch}{1.0} 
\centering
\resizebox{0.9\textwidth}{!}{
\begin{tabular}{llll}
\toprule
\textbf{ODE-guided generative models} & \textbf{Model} & \textbf{Formulation} &\textbf{Year}\\
\midrule
\multirow{9}{*}{\textbf{\makecell[l]{Flow-based models\\(Deterministic)}}} & Flow matching~\cite{lipman2022flow} & $dx(t)=v_\theta(x,t)\,dt$&2022\\
\cmidrule(lr){2-4}
& CFM ~\cite{lipman2022flow} & $    dx(t)=v_\theta(x,t)\,dt,\,x(1)\sim p_{\text{target}}(x|y)$ &2022\\
\cmidrule(lr){2-4}
& RFM~\cite{chen2023flow}& $dx(t)= v^{\mathcal{M}}_\theta(x,t)\,dt,\,x\in\mathcal{M}$&2023\\
\cmidrule(lr){2-4}
&DFM~\cite{gat2024discrete} &$dp(t) = Q_\theta(t)^\top p(t)\,dt$&2024\\
\cmidrule(lr){2-4}
&Fisher-Flow~\cite{davis2024fisher}&$dp(t) = v_\theta(p,t)\,dt,\,v_\theta(p,t)=g_p^{-1}\nabla_p\mathcal{L}(p,t)$&2024\\
\cmidrule(lr){2-4}
&SFM~\cite{cheng2024categorical}&$d\mu=v_\theta(\mu,t)\,dt,\mu\in\Delta^{n-1}$&2024\\
\cmidrule(lr){2-4}
&Generator Matching~\cite{holderrieth2025generator}&$dp(t)=\mathcal{G}_\theta^*(p,t)\,dt,\,\mathcal{G}_\theta^*\text{ is a generalized generating operator}$&2025\\
\cmidrule(lr){2-4}
&CS-DFM~\cite{cheng2025alpha}&$dx(t)=v_\theta(x,t)\,dt, x(t)=\pi^{(\alpha)}(\mu)$&2025\\
\cmidrule(lr){2-4}
&DiffoCFM~\cite{collas2025riemannian}&$dz(t)=v_\theta^E(z,t,y)\,dt,\,dx(t)=v_\theta^\mathcal{M}(x,t,y)\,dt,\,x(t)=\varphi^{-1}(z)$&2025\\
\midrule
\multirow{6}{*}{\makecell[l]{\textbf{Trajectory Optimization}\\\textbf{(Deterministic)}}} &Rectified Flow ~\cite{liu2022rectified,liu2023flow} &$  dx(t) = (x(1) -x(0))\,dt$&2022\\
\cmidrule(lr){2-4}
&InstaFlow~\cite{liu2023instaflow}&$x(1)=x(0)+v(x(0))\,T$&2023\\
\cmidrule(lr){2-4}
&PeRFlow~\cite{yan2024perflow} &$dx(t) = \left(x(1)^{(k)}-x(0)^{(k)}\right)\,dt,\,t\in[t_{k-1},t_{k}]$&2024\\
\cmidrule(lr){2-4}
&Hierarchical Rectified Flow~\cite{zhang2025towards}&$dx^{(l)}(t)=\left(x(1)^{(l)}-x(0)^{(l)}\right)\,dt,\,l=1,2,\cdots,L$&2025\\
\cmidrule(lr){2-4}
& MeanFlow~\cite{geng2026mean} &$x(1)=x(0)+\int_0^1 v_\theta(x,t)\,dt\approx x(0)+\overline{v}(x(0))$&2025\\
\cmidrule(lr){2-4}
&SlimFlow~\cite{zhu2025slimflow}&$x(1)=x(0)+v_\theta(x(0))$&2025\\
\toprule
\textbf{SDE-guided generative models} & \textbf{Model} & \textbf{Formulation} &\textbf{Year}\\
\midrule
\multirow{14}{*}{\textbf{\makecell[l]{Forward SDE\\(Stochastic)}}} 
&\textbf{Additive SDE}&$dx=f(x,t)dt+g(t)\,d\mathcal{W}(t)$&\\
\cmidrule(lr){2-4}
&NCSN~\cite{song2019generative} & $dx = \sqrt{\frac{d(g(t)^2)}{dt}}\,d\mathcal{W}(t)$& 2019\\
\cmidrule(lr){2-4}
&DDPMs~\cite{ho2020denoising} &$dx = -\frac12 \beta(t)x\,dt + \sqrt{\beta(t)}\,d\mathcal{W}(t)$ &  2020 \\
\cmidrule(lr){2-4}
&MSDiff~\cite{batzolis2021conditional} & 
\makecell[l]{
 $dx=\mu(x,t)dt+g_x(t)\,d\mathcal{W}(t)$\\
 $dy=\mu(y,t)dt+g_y(t)\,d\mathcal{W}(t)$\\
}
&2021\\
\cmidrule(lr){2-4}
&MRSDE~\cite{luo2023image}& $dx = \theta(t)(\mu-x)\,dt+g(t) \,d\mathcal{W}(t)$ &2023\\
\cmidrule(lr){2-4}
&BFNs~\cite{nie2023blessing}&$dx=xf(t)\,dt+g(t)\,d\mathcal{W}(t)$&2023\\
\cmidrule(lr){2-4}
& \makecell[l]{$\mathrm{L}\acute{\mathrm{e}}\mathrm{vy-It}\bar{\mathrm{o}}$ Model\\(LIM)~\cite{yoon2023score}}&$dx=f(x,t)dt+g(t)dL(t)$&2023\\
\cmidrule(lr){2-4}
&OU diffusion~\cite{cao2025generative} &$dx=-x(t)dt+\sqrt{2}d\mathcal{W}(t)$ &2024\\
\cmidrule(lr){2-4}
&\textbf{Multiplicative SDE}&$dx=f(x,t)dt+g(x,t)\,d\mathcal{W}(t)$&\\
\cmidrule(lr){2-4}
&SDMs~\cite{richemond2022categorical}&$dY^i = b(\alpha_i-Y^i)\,dt+\sqrt{2bY^i}d\mathcal{W}^i(t),\, x=Y/\sum_j Y^j$ &2022\\
\cmidrule(lr){2-4}
&FP-Diffusion~\cite{du2023flexible} & 
\makecell[l]{
$dx=f(x)dt+\sqrt{2R(x)}\,d\mathcal{W}(t)$\\
$dx = f(x)\beta'(t)\,dt+\sqrt{2\beta'(t)R(x)}\,d\mathcal{W}(t)$
}&2023\\
\midrule
\multirow{6}{*}{\makecell[l]{\textbf{Reverse SDE}\\\textbf{(Stochastic)}}}
 & \textbf{General reverse SDE}&$dx=[f(x,t)-g(t)^2\nabla_x\log\,p(x,t)]dt+g(t)\,d\mathcal{W}(t)$&\\
\cmidrule{2-4}
&DiffFlow~\cite{zhang2021diffusion} &\makecell[l]{
$dx=\big[f(x,t)+\beta(t)\nabla\,\log\frac{q(u(t)x,\sigma(t))}{p(x,t)}+\frac{g^2(t)}{2}\nabla\,\log p(x,t)\big]dt$\\
$\quad\quad+\sqrt{g^2(t)-\lambda^2(t)}d\overline{\mathcal{W}}(t)$}&2021\\
\cmidrule(lr){2-4}
&Unified Framework~\cite{cao2023exploring}&$dx=\left[f(x,t)-\gamma\cdot\frac12 g^2(t) s_\theta(x,t) \right]dt+\sqrt{1-\gamma^2}g(t)d\overline{\mathcal{W}}(t)$&2023\\
\cmidrule(lr){2-4}
& DEIS~\cite{zhang2022fast}&$dx=\left[f(t)x-\frac{1+\lambda^2}{2}g^2(t)s_\theta(x,t) \right]dt+\lambda g(t)d\overline{\mathcal{W}}(t)$&2023\\
\cmidrule(lr){2-4}
 &ER SDE~\cite{cui2025elucidating} & $ dx=\left[ f(t)x-\frac{g^2(t)+h^2(t)}{2}s_\theta(x,t)\right]dt+g(t)d\overline{\mathcal{W}}(t)$&2025\\
\midrule
\multirow{2}{*}{\makecell[l]{\textbf{Solvers of Reverse} \\\textbf{SDE (Stochastic)}}}& PF ODE~\cite{song2020score}&$dx=\left[f(x,t)-\frac12 g^2(t)s_\theta(x,t)\right]dt$&2020\\
\cmidrule(lr){2-4}
&DDIM~\cite{song2020denoising} &$dx=-\frac12\beta(t)[x+s_\theta(x,t)]dt$&2020\\
\bottomrule
\end{tabular}
}
\end{table*}

\subsection{Networks for Learning Deterministic Equations}\label{Sec:modeling_DD}
As illustrated in Fig.~\ref{fig:dl}, we partition deterministic architectures designed to learn or solve continuous equations into two primary domains. The first encompasses dynamics modeling via Neural DEs~\cite{chen2018neural} and ODE-guided generative models~\cite{lipman2022flow}, while the second focuses on Neural PDE solvers. Neural DEs~\cite{chen2018neural} are particularly effective for modeling sequential and continuous-time data, as they learn the governing dynamics $f_\theta(x,t)$ through neural network parameterization. In contrast, ODE-guided
generative models~\cite{lipman2022flow} focus on learning the velocity field $v_\theta(x, t)$ in the Liouville equation via neural networks, thereby enabling direct control over the evolution of probability densities. When solving equations, Neural PDE solvers~\cite{raissi2019physics,kovachki2023neural,lu2021learning} leverage deep learning to compute numerical solutions for complex physical systems by enforcing specific equation constraints. The details and taxonomy of Neural DEs, representative formulations and recent advances in ODE-guided generative models, and the core architectures of Neural PDE solvers are summarized in \cref{tab:De_dN}, \cref{tab:flow_model} and \cref{tab:neural_solver}, respectively.

\subsubsection{\textbf{Neural Deterministic Differential Equations}}
Integrating deterministic differential equations as continuous-depth architectural backbones provides a powerful deep learning paradigm for enhancing spatiotemporal physical simulations and complex sequential representation learning.

\textbf{Neural ODEs}~\cite{chen2018neural} transcend the rigid, discrete layers of traditional ResNets. By modeling continuous-time dynamics, they enable adaptive computation and formulate hidden state evolution via an ODE:
\begin{equation}
    \frac{dx(t)}{dt} =f_\theta(x,t),
\end{equation}
where $f_\theta$ parameterizes the continuous vector field. This formulation replaces discrete layers and solves the hidden state trajectory via an ODE, guaranteeing memory-efficient training through the adjoint method~\cite{pontryagin2018mathematical}. The continuous-time paradigm offers advantages such as constant memory cost regardless of model depth, suitability for irregularly sampled time series~\cite{chen2018neural,rubanova2019latent}, and adaptive computational depth. However, Neural ODEs come with drawbacks, including high computational cost during training and inference, as well as challenges with numerical instability and sensitivity to integration dynamics~\cite{kelly2020learning,chen2018neural,dupont2019augmented,norcliffe2021neural}.

\textbf{Second-Order Neural ODEs} (SONODEs) \cite{massaroli2020dissecting,norcliffe2020second} capture second-order dynamics by incorporating initial velocity and acceleration. This physics-inspired inductive bias yields more faithful representations of physical systems, enhancing both interpretability and numerical stability. Specifically, we consider SONODEs whose initial position $x(t_0)$, initial velocity $\dot{x}(t_0)$, and acceleration $\ddot{x}$ are given by:
\begin{equation*}
\ddot{x}=f_\theta(x,\dot{x},t),\,\dot{x}(t_0)=g_\theta(x(t_0)),\,x(t_0)=X_0,  
\end{equation*}
where $g_\theta$ and $f_\theta$ are neural networks predicting the initial velocity and acceleration. While convertible to a first-order ODE for training, this second-order formulation demonstrates superior physical interpretability, robustness, and extrapolation over ANODEs. High computational cost during training and inference remains a key challenge for Neural ODEs. Consequently, a range of methods have been proposed to improve their efficiency~\cite{rodriguez2022lyanet,mccallum2024efficient,bilovs2021neural,finlay2020train,ghosh2020steer,dupont2019augmented,liu2021second,nguyen2022improving,kelly2020learning,zhao2025accelerating,norcliffe2020second,lehtimaki2022accelerating,xia2021heavy}.
In parallel, to enhance the stability and robustness of Neural ODEs under long-term integration and noisy inputs, several approaches have been developed~\cite{tuor2020constrained,yan2019robustness,linot2023stabilized,kang2021stable,cui2023robustness,de2025stability,choromanski2020ode,massaroli2020stable,liu2019neural,zhang2025semi}.

Building upon the success of SONODEs, continuous deep architectures have naturally extended to spatiotemporal systems governed by PDEs. Shifting from traditional black-box dynamics fitting to white-box PDE discovery, \textbf{NeuPDE}~\cite{sun2020neupde} incorporates strong physical inductive biases. Combined with sparse regularization, it explicitly identifies the underlying governing equations:
\begin{equation}
   \frac{\partial u(t, x)}{\partial t} = \mathcal{N}_\theta (\mathcal{D}(u)) \approx \sum_{k=1}^{K} c_k \phi_k(u, \nabla u, \nabla^2 u, \dots), 
\end{equation}
where $c_k$ represent learnable sparse coefficients and $\phi_k$ denotes a library of candidate functions, comprising spatial derivatives and their nonlinear products. This formulation rigorously preserves physical consistency and enables direct mathematical extraction. Driven by a similar philosophy, PDE-Net~\cite{long2018pde,long2019pde} establishes mathematical mappings via constrained filter moment matrices to learn differential operators, achieving robust long-term forecasting and discovery of hidden PDE structures.

Standard Neural ODEs are limited in modeling irregularly sampled time series and in capturing complex temporal dynamics. Thus, researchers have developed several important extensions. Among these, \textbf{Neural Controlled Differential Equations} (Neural CDEs)~\cite{kidger2020neural} formulate the hidden state evolution as a controlled differential equation, providing:
\begin{equation*}
    z_t = z_{t_0} + \int_{t_0}^t f_\theta (z_s)\,dX_s, \,t\in(t_0,t_n),
\end{equation*}
where $z(t)$ denotes the hidden state, $X_s$ is an interpolated input path and $f_\theta$ controls how the state responds to input variations. Neural CDEs enable learning from irregularly sampled data by treating inputs as continuous trajectories. \textbf{Neural Rough Differential Equations} (Neural RDEs)~\cite{morrill2021neural} extend these methods to rough or non-differentiable paths using rough path theory~\cite{friz2020course}, capturing high-frequency or noisy dynamics through higher-order integration terms. \textbf{Neural Fractional Differential Equations} (Neural FDEs)~\cite{coelho2025neural} incorporate long-range memory and non-local dependencies by modeling
\begin{equation*}
    D^\alpha z(t) = f_\theta(z(t),t),\,0<\alpha<1,
\end{equation*}
where $D^\alpha$ is a Caputo or Riemann–Liouville fractional derivative. Neural FDEs consider history-dependent dynamics, making them suitable for non-Markovian systems.

As summarized in \cref{tab:De_dN}, continuous architectures evolved to overcome the theoretical and numerical limits of early deterministic models. While standard Neural ODEs~\cite{chen2018neural} pioneered continuous-depth modeling, their topology-preserving nature and Markovian assumptions restrict physical realism and induce severe numerical stiffness. Architectures resolve this along two trajectories. The first injects physical inductive biases to create interpretable models, utilizing augmented state spaces in ANODEs~\cite{dupont2019augmented} and second-order fields in SONODEs~\cite{norcliffe2020second}. The second relaxes autonomous constraints to capture history-dependent dynamics via control paths in Neural CDEs~\cite{kidger2020neural} or fractional derivatives in Neural FDEs~\cite{coelho2025neural}. Ultimately, enhancing physical interpretability and memory shifts the computational bottleneck toward complex numerical integration, demanding strict alignment between the DE framework and computational resources.

\subsubsection{\textbf{ODE-Guided Generative Models}}
In generative modeling, deterministic differential equations serve as optimal transport frameworks that smoothly map simple base distributions to complex target data manifolds.

ODE-guided Generative Models~\cite{lipman2022flow,lipman2024flow} deterministically transport a simple base distribution to a complex target distribution. This transformation is achieved by learning a time-dependent velocity field $v_\theta(x,t)$ that governs the continuous evolution of data points via an ODE:
\begin{equation}
    \frac{dx}{dt} = v_\theta(x,t),\,x(0)\sim p(0),\,x(1)\sim p_{\text{target}} (x).
\end{equation}
Under this deterministic flow, probability density evolution follows the Liouville equation. Flow Matching~\cite{lipman2022flow} introduces a simulation-free objective to overcome the expensive ODE simulations traditionally required to track this density. It constructs a target velocity field via conditional path interpolation and trains $v_\theta$ to regress it directly, thereby completely bypassing explicit density computation and enabling scalable high-dimensional training.

Standard flow matching is restricted to continuous Euclidean spaces~\cite{chen2023flow} and linear trajectories~\cite{liu2023flow,albergo2023building}. To overcome these limitations, subsequent research advances along two primary directions: extending the framework to non-Euclidean topologies, such as discrete domains~\cite{campbell2022continuous,gat2024discrete}, graphs~\cite{hou2024improving}, and manifolds~\cite{10.5555/3737916.3742207,chen2023flow}, and optimizing transport paths~\cite{liu2022rectified,liu2023flow} to accelerate sampling, ultimately enabling one-step generation~\cite{geng2026mean}.

\textbf{Generalized Flow Matching}: 
A series of generalized flow matching approaches have emerged, each systematically extending the framework toward broader classes of data and structural priors. Notable examples include:

\begin{itemize}
    \item \textbf{ Conditional Flow Matching (CFM)}~\cite{lipman2022flow} is a method that learns a velocity field \(v_\theta(x, t, y)\), which is conditioned on an auxiliary variable \(y\), allowing for controllable generation. The system is governed by the following ODE:
    \[
\frac{dx}{dt} = v_\theta(x, t),\,  x(0) \sim p(0), \, x(1) \sim p_{\text{target}}(x|y),
\]
which facilitates the direct generation of samples with specified attributes to ensure alignment with conditioning information. This capability is particularly beneficial for tasks such as class-conditional image synthesis.

\item \textbf{Discrete Flow Matching (DFM)} \cite{campbell2022continuous,gat2024discrete} generalizes flow matching to discrete and categorical data domains by replacing the continuous flow with a continuous-time Markov jump process. The dynamics are governed by a time-dependent rate matrix $Q_\theta(t)$:
\[
\frac{dp(t)}{dt} = Q_\theta(t)^\top p(t),
\]
where $p(t)$ denotes the time-evolving probability mass over discrete states. This approach enables the efficient, non-autoregressive generation of sequence, graph, and molecular data.
\end{itemize}

These developments mark a shift in flow matching research toward modeling data in non-Euclidean~\cite{10.5555/3737916.3742207,chen2023flow} and structured spaces~\cite{cheng2024categorical,davis2024fisher,cheng2025alpha}. Riemannian Flow Matching (RFM) \cite{chen2023flow} adapts flow matching to non-Euclidean domains by respecting manifold constraints, enabling generative modeling on curved data spaces. Generator Matching \cite{holderrieth2025generator} generalizes flow matching to arbitrary modalities and Continuous Time Markov Processes (CTMPs), supporting non-autoregressive generation for discrete sequences and graphs. Methods like Continuous-State Discrete Flow Matching (CS-DFM)~\cite{cheng2025alpha}, Fisher-Flow~\cite{davis2024fisher}, and Statistical Flow Matching (SFM)~\cite{cheng2024categorical} bring in information geometry, aligning probability transport with natural statistical metrics such as the Fisher–Rao distance. Diffeomorphism Conditional Flow Matching (DiffeoCFM)~\cite{collas2025riemannian} further extends flow matching to positive-definite matrix manifolds via pullback geometry. Altogether, these innovations substantially broaden the applicability and expressivity of flow matching, providing a robust geometric foundation for ODE-guided generative modeling in complex domains.

\textbf{Trajectory Optimization:}
Trajectory Optimization explores advanced strategies to optimize both the generative trajectory and its numerical integration, aiming to improve sampling efficiency, stability, and accuracy~\cite{dao2023flow,wang2025training,DBLP:journals/corr/abs-2407-02398}. 

\begin{itemize}
    \item \textbf{Rectified Flow} \cite{liu2022rectified,liu2023flow} simplifies the generative process by constraining the transport path between noise and data to a straight line in data space, allowing for highly efficient and deterministic sampling. The underlying ODE is
\[\frac{dx}{dt} = x(1) -x(0),\]
where $x(0)$ and $x(1)$ are the endpoints of the trajectory, allowing the model to learn a velocity field that maps base samples directly along a rectified path.  

\item \textbf{MeanFlow} \cite{geng2026mean,geng2025improved} adopts a one-step generation approach by learning a time-averaged velocity field \(\overline{v}_\theta(x, t)\). This enables the model to directly map a base sample to the target in a single step:
\[
x(1) = x(0) + \int_0^1 v_\theta(x, t)\, dt \approx x(0) + \overline{v}(x(0)).
\]
\end{itemize}

Within this trajectory-optimization perspective, subsequent studies mainly revisit three closely related design issues: how to train the transport dynamics more robustly, how to choose more expressive transport paths, and how to reduce the cost of numerical sampling. Work on training objectives and optimization dynamics addresses issues such as velocity variance and trajectory crossings, which can affect training stability and sampling efficiency~\cite{lee2024improving,NEURIPS2024_a3f01962,Schusterbauer_2025_CVPR,ma2025flow,guo2025variational}. Path-oriented studies move beyond a single global straight-line transport by introducing hierarchical, piecewise, or more flexible trajectories, allowing the generative path to better adapt to complex data geometries~\cite{zhang2025hierarchical,yan2024perflow,wang2024rectified}. Efficiency-oriented methods, including InstaFlow~\cite{liu2023instaflow}, Optimal Flow Matching~\cite{kornilov2024optimal}, and SlimFlow~\cite{zhu2025slimflow}, seek to reduce the number of function evaluations through distillation, shortcut mappings, or optimized transport objectives. Taken together, these studies show that trajectory optimization in flow-matching models has progressed from fixed straight-line transports toward more stable, adaptive, and computationally efficient transport formulations~\cite{zhu2024flowie,li2025omniflow,zhu2025slimflow}.

\subsubsection{\textbf{Neural PDE solvers}}
PDEs are fundamental for modeling physical and engineering systems. Although traditional numerical methods such as FEM and FVM are reliable, they often suffer from high computational cost in high-dimensional or parametric settings. Neural PDE solvers address this limitation by using DNNs to approximate specific PDE solutions or the underlying solution operators. 

\textbf{Physics-Informed Neural Networks (PINNs).} PINNs~\cite{yu2018deep,raissi2019physics,cuomo2022scientific,cai2021physics} incorporate governing equations into network training to approximate either specific PDE solutions. For a PDE $\mathcal{N}[u](x,t)=0$, the surrogate solution $u_{\theta}$ is optimized by minimizing
\begin{equation}\label{eq:loss}
\mathcal{L}
=
\omega_{\text{pde}}\mathcal{L}_{\text{pde}}
+
\omega_{\text{bound}}\mathcal{L}_{\text{bound}}
+
\omega_{\text{init}}\mathcal{L}_{\text{init}}
\quad\mbox{with}~~
\mathcal{L}_{\text{pde}}
=
\frac{1}{N_r}\sum_{i=1}^{N_r}
\left|
\mathcal{N}[u_{\theta}](x_i,t_i)
\right|^2.
\end{equation}
The main challenge lies in optimization instability caused by PDE stiffness and imbalance among loss terms. Current improvements mainly follow two directions. 
\textbf{(1) Loss balancing strategies:} adaptive weights~\cite{wang2022and} are introduced using NTK theory~\cite{jacot2018neural}, self-adaptive mechanisms~\cite{xiang2022self,mcclenny2023self}, and gradient-enhanced methods~\cite{yu2022gradient,son2021sobolev}. 
\textbf{(2) Variational and energy-based formulations:} weak-form PINNs such as VPINNs~\cite{kharazmi2019variational} minimize
\begin{equation}\label{eq:variation_loss}
\mathcal{L}_{\text{pde}}
=
\frac{1}{K}\sum_{i=1}^{K}
\left|
\int_{\Omega}
\varphi_i(x)\mathcal{N}[u_{\theta}](x)\,dx
\right|^2,
\end{equation}
and are further extended by local test functions~\cite{kharazmi2021hp}, adversarial weak forms~\cite{zang2020weak}, and energy minimization~\cite{samaniego2020energy}.

\textbf{Operator Learning.} Unlike PINNs, which approximate solutions for specific instances, operator learning aims to learn the mapping $\mathcal{G}: \mathcal{F} \to \mathcal{U}$ between infinite-dimensional function spaces $\mathcal{F}$ and $\mathcal{U}$~\cite{kovachki2023neural}. Given $N$ input-output function pairs and $P$ query points per sample, the model is typically optimized via the empirical supervised loss:
\begin{equation}
\mathcal{L}_{\text{data}}
=
\frac{1}{NP}
\sum_{i=1}^{N}\sum_{j=1}^{P}
\left|
\mathcal{G}_{\theta}(f^{(i)})(y_j^{(i)})
-
\mathcal{G}(f^{(i)})(y_j^{(i)})
\right|^2.
\end{equation}

\textbf{DeepONet}~\cite{lu2021learning} leverages the universal approximation theorem for operators by decomposing the mapping into branch and trunk networks. The operator is represented as a generalized basis expansion:
\begin{equation}\label{eq:basis_epansion}
\mathcal{G}_{\theta}(f)(y)
=
\sum_{k=1}^{p}
b_k\big(f(x_1),\dots,f(x_m)\big)t_k(y)+b_0.
\end{equation}
Notable extensions include POD-DeepONet~\cite{lu2022comprehensive} for reduced-order modeling, GreenONet~\cite{aldirany2024operator} based on Green's functions, and variants tailored for geometry or sequential data such as Geom-DeepONet~\cite{he2024geom} and Sequential DeepONet~\cite{he2024sequential}.

The \textbf{Fourier Neural Operator} (FNO)~\cite{li2020fourier} parameterizes the integral kernel in the spectral domain, enabling the capture of global dependencies. The architecture updates the hidden state $v_k$ through a combination of a linear transform and a spectral convolution:
\begin{equation}
v_{k+1}(y) = \sigma \left( W v_k(y) + \mathcal{F}^{-1} \big( R_k \cdot \mathcal{F}(v_k) \big) (y) \right),
\end{equation}
where $\mathcal{F}$ denotes the Fourier transform and $R_k$ is a learnable weight matrix in the frequency domain. This framework has been generalized to other functional bases, including wavelets, polynomials, and Laplace transforms~\cite{tripura2023wavelet,fanaskov2023spectral,cao2024laplace,qin2026starter}.

\textbf{Physics-Informed Operator Learning} \cite{wang2021learning} integrates operator learning with the PDE residual as a regularization ter,m to enforce physical constraints
\begin{equation}\label{eq:pino_loss}
\mathcal{L}
=
\mathcal{L}_{\text{data}}
+
\mathcal{L}_{\text{pde}},
\end{equation}
where $\mathcal{L}_{\text{pde}}$ penalizes violations of the underlying governing equations. This paradigm enhances physical consistency and generalization in data-scarce regimes. Subsequent developments include PINO~\cite{li2024physics}, PIWNO~\cite{navaneeth2024physics}, and multi-resolution frameworks~\cite{goswami2023physics,roy2025physics}.

Overall, Neural PDE solvers have evolved from pointwise residual learning to operator-level approximation, becoming a core direction in scientific machine learning. PINNs require little labeled data and preserve physical interpretability, but often suffer from optimization difficulties and poor scalability. Operator learning enables rapid inference and strong generalization across parameter spaces, but usually depends on large high-quality datasets. Physics-informed operator learning provides a promising compromise by combining data efficiency with physical consistency. A detailed summary is provided in \cref{tab:neural_solver}. Furthermore, while the above methods assume a fully known governing PDE, Universal Differential Equations (UDEs)~\cite{rackauckas2020universal} have been developed for scenarios where the PDE is only partially known, learning the unknown terms by embedding trainable sub-networks directly into the solution procedure. The rapid development of these solvers is driven by the growing demand for fast, scalable, and reliable surrogate models in computational science, especially for high-dimensional PDEs and multi-query simulations.

\begin{table*}[!t]
\caption{Summary of neural network solvers for numerical PDEs: Physics-informed methods and operator learning methods.}
\label{tab:neural_solver}
\centering
\renewcommand{\arraystretch}{1.0} 
\resizebox{0.95\textwidth}{!}{
\begin{tabular}{llll}
\toprule
\textbf{Category} & \textbf{Model} & \textbf{Core Formulation / Loss Objective} & \textbf{Year} \\
\midrule
\multirow{9}{*}
{\textbf{PINNs}} & Deep Ritz \cite{yu2018deep} & Energy-based minimization of the Ritz functional $I(u) = \int_{\Omega} f(\nabla u, u) dx$ & 2018\\
\cmidrule{2-4}
& Vanilla PINNs \cite{raissi2019physics} & $\mathcal{L}(\theta) = \mathcal{L}_{\text{pde}} + \mathcal{L}_{\text{bound}} + \mathcal{L}_{\text{init}}$ via automatic differentiation & 2019 \\
\cmidrule{2-4}
& VPINNs \cite{kharazmi2019variational} & Variational formulation using global Petrov-Galerkin test functions & 2019\\
\cmidrule{2-4}
& WAN \cite{zang2020weak} & Weak form residual via adversarial training: $\min_{u} \max_{v} \mathcal{L}_{\text{weak}}(u, v)$ & 2020\\
\cmidrule{2-4}
& hp-VPINNs \cite{kharazmi2021hp} & Domain decomposition with local test functions: $\mathcal{R} = \langle \mathcal{N}[u_\theta], v \rangle_{\Omega_e}$ & 2021\\
\cmidrule{2-4}
& Loss reweighting \cite{wang2022and} & NTK-based weighting: $\mathcal{L}(\theta) = \sum \omega_i \mathcal{L}_i$ using eigenvalues of the NTK matrix & 2022 \\
\cmidrule{2-4}
& lbPINNs \cite{xiang2022self} & Likelihood-based adaptive weights: $\sum \frac{1}{2\varepsilon^2_i} \mathcal{L}_i + \log \prod \varepsilon_i$ & 2022\\
\cmidrule{2-4}
& gPINNs \cite{yu2022gradient} & Gradient-enhanced residual: $\mathcal{L}_{\text{total}} = \mathcal{L}_{\text{pde}} + \omega_{\text{grad}}\mathcal{L}_{\nabla \text{pde}}$ & 2022\\
\cmidrule{2-4}
& SAPINNs \cite{mcclenny2023self} & Mask-based trainable weights: $\mathcal{L} = \sum \mathcal{L}_i(\theta, \omega_i)$ via mini-max optimization & 2023\\
\midrule
\multirow{10}{*}{\textbf{Neural Operator}} & FNO \cite{li2020fourier} & Spectral convolution via FFT: $\mathcal{K}(v) = \mathcal{F}^{-1}(R \cdot \mathcal{F}(v))$ & 2020 \\
\cmidrule{2-4}
& DeepONet \cite{lu2021learning} & Operator mapping via basis expansion: $G(u)(y) \approx \sum_{k=1}^q b_k(u) \tau_k(y)$ & 2021 \\
\cmidrule{2-4}
& PI-DeepONet \cite{wang2021learning} & Physics-regularized DeepONet: $\mathcal{L}_{\text{data}} + \lambda \mathcal{L}_{\text{pde}}$ & 2021\\
\cmidrule{2-4}
& POD-DeepONet \cite{lu2022comprehensive} & POD-derived basis functions replacing the trunk net $\tau_k(y)$ & 2022 \\
\cmidrule{2-4}
& WNO \cite{tripura2023wavelet} & Multiresolution operator mapping via Wavelet Transform decomposition & 2023\\
\cmidrule{2-4}
& SNO \cite{fanaskov2023spectral} & Spectral mapping using orthogonal polynomial bases (e.g., Chebyshev) & 2023\\
\cmidrule{2-4}
& LNO \cite{cao2024laplace} & Exponential basis expansion via Laplace Transform for transient dynamics & 2024\\
\cmidrule{2-4}
& PINO \cite{li2024physics} & Physics-informed FNO: Integrating $\mathcal{L}_{\text{pde}}$ into the spectral architecture & 2024\\
\cmidrule{2-4}
& PIWNO \cite{navaneeth2024physics} & Wavelet-based operator learning constrained by PDE residuals & 2024\\
\cmidrule{2-4}
& VINO \cite{eshaghi2025variational} & Variational operator learning: FNO architecture with weak-form loss & 2025\\
\bottomrule
\end{tabular}
}
\end{table*}

\section{From SDEs to Deep Neural Networks}
Incorporating stochasticity into ODEs yields SDEs, providing a more expressive framework for modeling uncertainty. These equations typically feature additive noise to account for external disturbances, or multiplicative noise to capture state-dependent randomness like thermal variations. The integration of SDEs with neural networks has established a new machine learning paradigm, enabling powerful regularization techniques and the principled modeling of stochastic dynamics. This intersection primarily manifests in two domains:
\begin{itemize}
    \item \textbf{Stochastic Network Regularization (Networks Guided by Equations)}: Controlled noise is injected at various network levels to enhance robustness and generalization. This includes neuron-level Dropout \cite{baldi2013understanding,srivastava2014dropout}, weight perturbations \cite{lian2016dropconnect,kang2017shakeout}, activation noise \cite{moradi2020survey,li2016whiteout,he2019parametric}, and dynamic depth adjustment \cite{gastaldi2017shake,huang2016deep}. By mimicking stochastic perturbations in the optimization landscape, these methods help avoid poor local minima.
    
    \item \textbf{Stochastic Dynamics Modeling}: Forward propagation in neural networks can be mathematically characterized as a dynamical system driven by SDEs. Neural SDEs~\cite{liu2019neural} parameterize both drift and diffusion coefficients to model complex stochastic dynamics. In contrast, SDE-guided generative models (e.g., DDPMs)~\cite{sohl2015deep,ho2020denoising} define a fixed forward noise process and generate data by learning the drift term of the reverse SDE. Fundamentally, they differ in their learning objectives: Neural SDEs capture forward-time stochastic dynamics, whereas DDPMs learn reverse-time denoising dynamics.
\end{itemize}

\subsection{Stochastic Network regularization}\label{Sec:Network_regularition}

We begin by reviewing stochastic regularization methods that introduce randomness into different components of neural networks, such as neurons, weights, activations, and architectural paths. Representative methods are summarized in \cref{tab:stochastic_regularizaztion}, along with their key design choices and regularization effects.

\begin{table*}[!t]
\caption{Summary of SDE-guided stochastic regularization methods for deep neural network models, categorized by their application to neurons, weights, activations, and layers.}
\label{tab:stochastic_regularizaztion}
\centering
\resizebox{\textwidth}{!}{
\begin{tabular}{llll}
\toprule
\textbf{Category} & \textbf{Model} & \textbf{Formulation} & \textbf{Year} \\
\midrule
\multirow{7}{*}{\textbf{Neurons}}  & Dropout \cite{hinton2012improving,srivastava2014dropout} &$x_{n+1}=x_n+f(x_n,\omega_n)\odot\frac{z_n}{p}, ~z_n\sim\text{Bernoulli}(p)$ & 2012 \\
\cmidrule(lr){2-4}
 & Fast Dropout \cite{wang2013fast} &$x_{n+1}=(1-p)x_n+\mathcal{N}(0,p(1-p)x_n^2)$ &2013\\
 \cmidrule(lr){2-4}
 &Annealed Dropout~\cite{rennie2014annealed}& $x_{n+1}=x_n+f(x_n,\omega_n)\odot\frac{z_n(t)}{p},~z_n\sim\text{Bernoulli}(p(t))~\mbox{and}~p(t)=p(t-1)+\alpha_t(\theta)$ &2014\\
 \cmidrule(lr){2-4}
 &SpatialDropout~\cite{tompson2015efficient} &$\tilde{y}_c^{l-1}=z_c\cdot y_c^{l-1},~\text{$y_c^{l-1}$ is the c-th channel feature map}$&2014\\
 \cmidrule(lr){2-4}
 &Curriculum Dropout~\cite{morerio2017curriculum}& $x_{n+1}=x_n+f(x_n,\omega_n)\odot\frac{z_n(t)}{p},\,z_n\sim\text{Bernoulli}(p(t))~\mbox{and}~p(t)=(1-\theta)(1-e^{-\gamma t})$ &2017\\
 \cmidrule(lr){2-4}
 & Targeted Dropout~\cite{gomez2019learning}& $x_{n+1}=x_n+f(x_n,\omega_n)\odot\frac{z_n}{p},\,z_n\sim\text{Bernoulli}(p_n)~\mbox{and}~p_n=\text{TargetDropout}(s_n) $  &2018\\
 \cmidrule(lr){2-4}
  & DropBlock~\cite{ghiasi2018dropblock} & $\tilde{A}=A\odot M,\, M_{i,j}=\text{Bernoulli}(\gamma)~\mbox{and}~\text{$A$ is the feature map}$  & 2018 \\
  \midrule
\multirow{3}{*}{\textbf{Weights}} & DropConnect~\cite{wan2013regularization} &  $\tilde{W}=W\odot M,\,M_{i,j}=\text{Bernoulli(1-p)}$ & 2013  \\
\cmidrule(lr){2-4}
 &Variational Dropout~\cite{kingma2015variational}& $W\sim\mathcal{N}(\mu,\alpha\mu^2),\,\alpha \,\text{is a learnable parameter.}$  &2015\\
 \cmidrule(lr){2-4}
 &Shakeout~\cite{kang2017shakeout}& $\tilde{W}_{ij}=c(r_j-1)S_{ij}-r_j\cdot(\frac1p W_{ij}+c\frac{1}{1-p}S_{ij}),S_{ij}=\text{sgn}(W_{ij}),r_j\sim\text{Bernoulli}(p)$ &2016\\
\midrule
\multirow{3}{*}{\textbf{Activations}}  &Whiteout\cite{li2016whiteout}& $x_{n+1}=x_n+f(x_n)+\epsilon,\,\epsilon\sim\mathcal{N}(0,\sigma^2|x|^{2\gamma})$ &2016\\
\cmidrule(lr){2-4}
&GNI~\cite{moradi2020survey}& $x_{n+1}=x_n+f(x_n)+\sigma\mathcal{N}(0,I)$ &2020\\
\cmidrule(lr){2-4}
    &PNI~\cite{he2019parametric}& $x_{n+1}=x_n+\alpha\cdot\mathcal{N}(0,I),\,\alpha \,\text{is a learnable parameter.}$ &2019\\
\midrule
\multirow{8}{*}{\textbf{Layers}} & Stochastic Depth~\cite{huang2016deep} & $x_{n+1}=x_n+z_n\,f(x_n),\, z_n\sim\text{Bernoulli}(p)$ & 2016 \\
\cmidrule(lr){2-4}
 &DropPath~\cite{larsson2016fractalnet} & $x_{n+1}=x_n+\sum\limits_i z_{in} f_i(x_n),\,z_{in}\sim\text{Bernoulli}(p)$ &2016\\
 \cmidrule(lr){2-4}
  &Swapout~\cite{singh2016swapout}&  $x_{n+1}=\alpha\,x_n+\beta\,f(x),\,\alpha,\beta\sim\text{Bernoulli}(p)$ &2016\\
  \cmidrule(lr){2-4}
 & Shake-shake regularization~\cite{gastaldi2017shake} & $x_{n+1}=x_n+\alpha_n f_1(x)+(1-\alpha_n)f_2(x),\,\alpha_n\sim\text{Bernoulli}(p)$ & 2017  \\
 \cmidrule(lr){2-4}
 &PyramidSepDrop~\cite{yamada2017deep}& $x_{n+1}=x_n+\sum\limits_{i}\alpha_i f_i(x),\,f(x)=[f_1(x),\cdots,f_K(x)],\,\alpha_i\sim\text{Bernoulli}(1-p)$ &2017\\
 \cmidrule(lr){2-4}
 &ShakeDrop~\cite{yamada2019shakedrop} & $x_{n+1}=x_n+\alpha_n\cdot b\cdot f(x),\,\alpha_n\sim\text{Bernoulli}(p),\,b\sim\text{U}([-1,1])$ &2018\\
 \cmidrule(lr){2-4}
 &Scheduled drophead~\cite{zhou2020scheduled}& $\text{head}_i^{\text{dropped}}=\alpha_i(t)\cdot\text{head}_i,\, \alpha_i(t)\sim\text{Bernoulli}(1-p),\,p(t)=p_{\max}\cdot\frac{t}{T}$ &2018\\
 \cmidrule(lr){2-4}
 &LayerDrop~\cite{fan2019reducing} & $h_{l+1}=\alpha_l\cdot\text{Layer}_l(h_l)+(1-\alpha_l)\cdot h_l,\,\alpha_l\sim\text{Bernoulli}(1-p)$ &2019\\
\bottomrule
\end{tabular}
}
\end{table*}

\subsubsection{\textbf{Dropout}~\cite{hinton2012improving,srivastava2014dropout}} Dropout is a fundamental regularization technique that randomly deactivates hidden units during training to mitigate overfitting. When integrated with ResNets, the forward propagation through a dropout-augmented residual block is formally expressed as:
\[x_{n+1} = x_n + f(x_n,\omega_n)\odot \frac{z_n}{p},~~z_n\sim\text{Bernoulli(p)},\]
where $p$ is the retention probability, $\odot$ denotes element-wise multiplication. Dropout has been theoretically interpreted as an approximation to Bayesian inference~\cite{ba2013adaptive}, with further formalization accomplished through a SDE~\cite{liu2020does,wang2025convection}, given by:
\begin{equation}
    dx(t) = f(x)dt + \sqrt{\frac{1-p}{p}} \,f(x)\cdot d\mathcal{W}(t),
\end{equation}
where $\mathcal{W}(t)$ has the same meaning as in SDEs. This perspective enables stability analysis and supports adaptive optimization.

Since the introduction of standard dropout, numerous neuron-level stochastic regularization variants have emerged. Fast Dropout~\cite{wang2013fast} leverages Gaussian approximation for efficiency, while Annealed~\cite{rennie2014annealed} and Curriculum Dropout~\cite{morerio2017curriculum} adjust noise levels over training. Variational Dropout~\cite{kingma2015variational} treats weights as Gaussian variables for adaptive regularization and sparsity. Other extensions include structure-aware methods such as SpatialDropout~\cite{tompson2015efficient}, Max-pooling and Convolutional Dropout~\cite{wu2015max}, as well as selective masking strategies like Targeted Dropout~\cite{gomez2019learning}, DropBlock~\cite{ghiasi2018dropblock}, and EDropout~\cite{salehinejad2021edropout}.

\subsubsection{\textbf{DropConnect}~\cite{wan2013regularization}} DropConnect regularizes neural networks by randomly masking weights rather than neurons. It applies a binary mask $M$, sampled as $M_{i,j} \sim \text{Bernoulli}(1-p)$, to the weight matrix $W$, producing an output:  
\begin{equation*}  
    y = f_a((W \odot M)\,x),\,x_{n+1}^i = \sum_j (W_{ij}\, x_{n}^j) M_{ij}
\end{equation*}  
where $\odot$ denotes element-wise multiplication and $f_a$ is a non-linear activation function. Theoretically, DropConnect reduces to Dropout in shallow networks \cite{pal2020regularization} and can be represented by the following SDE:  
\begin{equation*}  
   d x(t) = pW x\, dt + \sqrt{p(1 - p)}\, \left(W \odot W\right)^{1/2}   
    \left(x \odot x\right)^{1/2}d\mathcal{W}(t),
\end{equation*}  
which has inspired numerous extensions, including Sparsity-aware masking \cite{lian2016dropconnect},  Output uncertainty modeling via Dirichlet distributions \cite{zheng2022structured}, Gradient-based dynamic drop rates \cite{yang2025dynamic}, Adaptive drop rates using generalization gap and complexity measures \cite{joudal2022adaptive}.

\subsubsection{\textbf{Gaussian Noise Injection (GNI)}  \cite{moradi2020survey}} GNI enhances model generalization by perturbing intermediate representations during training. In a standard ResNet with $L$ residual blocks, the output of the $n$-th block with injected noise is given by:
\begin{equation*}
    x_{n+1} = x_n + f(x_n) + \alpha\,\mathcal{N}(0,I),
\end{equation*}
where $\alpha$ controls the noise intensity and $\mathcal{N}(0,I)$ is standard Gaussian noise. By setting $\alpha = \sigma \sqrt{\Delta t}$ and interpreting $f(x_n)$ as a discretized velocity field $v(x_n, t_n)$, this update rule corresponds to the Euler–Maruyama discretization of SDEs:
\begin{equation*}
    dx(t) = v(x,t)\,dt + \sigma\,d\mathcal{W}(t).
\end{equation*}
Whiteout~\cite{li2016whiteout} injects adaptive Gaussian noise into activations to promote structured sparsity, an approach later grounded in a variational interpretation~\cite{noh2017regularizing}. Theoretical analyses reveal that such activation-level noise penalizes high-frequency components~\cite{camuto2020explicit}, acts as ridge regularization in the infinite-sample limit~\cite{dhifallah2021inherent}, and benefits overparameterized regimes~\cite{orvieto2023explicit}. To enhance adversarial robustness, Parametric Noise Injection (PNI)~\cite{he2019parametric} applies learnable noise to activations or weights. Ultimately, Gaussian noise injection serves as a principled regularization mechanism that smooths the loss landscape and bridges discrete training with continuous-time stochastic dynamics~\cite{ye2023improving,orvieto2023explicit}.

\subsubsection{\textbf{Stochastic Depth}~\cite{huang2016deep}}
Stochastic Depth mitigates overfitting by randomly bypassing residual blocks during training. For an $L$-layer residual network, the forward propagation at the $n$-th block is formulated as:
\begin{equation*}
    x_{n+1} = x_n + z_n \,f(x_n),\quad z_n\sim\text{Bernoulli(p)},
\end{equation*}
where $p$ denotes the layer-wise survival probability. The network with stochastic drop can be written as an SDE
\begin{equation*}
    dx(t) = p f(x) dt + \sqrt{p(1 - p)} F(x) d\mathcal{W}(t).
\end{equation*}

Recent studies have explored the theoretical and empirical properties of stochastic depth~\cite{hayou2021regularization}. Building on foundational block-level stochasticity, subsequent research extends dropping mechanisms through fine-grained structural perturbations and dynamic probability scaling. Structurally, PyramidSepDrop \cite{yamada2017deep} independently drops channels, DropPath \cite{larsson2016fractalnet} discards specific residual paths, and Swapout \cite{singh2016swapout} perturbs both identity and residual branches. To balance regularization with training stability, survival probabilities are dynamically calibrated: Scheduled DropPath \cite{zhou2020scheduled} modulates rates over training time, while LayerPath \cite{fan2019reducing} scales them across network depth. Furthermore, ShakeDrop \cite{yamada2019shakedrop} introduces stochastic path blending to stabilize single-branch architectures. Collectively, these progressive refinements underscore the critical role of structured noise in robust network optimization and generalization.

\subsection{Stochastic Dynamics Modeling}\label{sec:dynamic_SDEs}
SDEs provide a powerful mathematical framework for modeling random dynamical systems in deep learning. Two major classes of SDE-driven approaches have emerged. The first, Neural SDEs~\cite{liu2019neural} utilize neural networks to parameterize both drift and diffusion terms, providing a flexible framework for modeling complex stochastic dynamics and effectively capturing temporal uncertainty in time-series analysis. A detailed categorization and summary of these methods can be found in \cref{tab:De_dN}. The second class consists of SDE-guide generative models, represented DDPMs~\cite{song2020score}. The evolution of probability densities is described by the FPE, and the sampling process can be expressed using the SDE. DDPMs learn the score or noise component to guide the evolution of probability densities, avoiding explicit density computation. Detailed comparisons are summarized in \cref{tab:flow_model}.

\subsubsection{\textbf{Neural SDEs}~\cite{liu2019neural}}  Neural SDEs extend Neural ODEs~\cite{chen2018neural} by introducing stochasticity through a learnable diffusion term, enabling the modeling of systems with intrinsic randomness. Unlike the deterministic trajectories of Neural ODEs, Neural SDEs capture uncertainty via stochastic processes parameterized by neural networks. The latent state $x_t$ evolves according to the It\^o SDE
\begin{equation}\label{SDE1}
    dx(t) = f_\theta(x,t)\,dt+g_\theta(x,t)\,d\mathcal{W}(t),
\end{equation}
where $f_\theta$ and $g_\theta$ denote the drift and stochastic driving networks, respectively. Eq.~\eqref{SDE1} is typically simulated via numerical methods like Euler–Maruyama~\cite{maruyama1955continuous}, with gradients computed through reparameterization or adjoint methods~\cite{pontryagin2018mathematical}. While Neural SDEs excel at modeling noisy or irregular data across time series and generative tasks, they suffer from significant computational overhead and limited interpretability.

Recent advancements in Neural SDEs can be categorized into four main areas. Firstly, they have been developed as expressive generative models: Tzen et al.~\cite{tzen2019neural} link them to diffusion-limit approximations of latent Gaussian models, Li et al.~\cite{li2020scalable} extend latent ODEs by integrating stochastic priors, and Kidger et al.~\cite{kidger2021neural} adapt both the generator and discriminator of GANs as Neural SDEs. Secondly, in terms of efficiency and stability, efforts have concentrated on enhancing solvers and objectives. Reversible solvers~\cite{kidger2021efficient} and deterministic approximations~\cite{look2022deterministic} have been employed to mitigate integration overhead, while alternative objectives such as signature kernel losses~\cite{issa2023non} and marginal-matching criteria~\cite{zhang2024efficient} bolster training robustness. Complementary improvements include cubature methods on Wiener space~\cite{snow2025efficient}, finite-dimensional distribution matching~\cite{zhang2025efficient}, stochastic optimal control formulations~\cite{01JRZ4VX2Z72Y7ZMP7XG8CHBX9}, and high-dimensional gradient estimators~\cite{NEURIPS2024_a0cd56b9}, all contributing to enhanced scalability and optimization efficiency. Thirdly, Neural SDEs have been applied across various fields such as robust financial modeling~\cite{gierjatowicz2020robust}, spatiotemporal dynamics~\cite{salvi2022neural}, and unified sequence modeling~\cite{shen2025neural}. Finally, recent studies have improved the modeling of rare and irregular dynamics: Neural Jump SDEs~\cite{jia2019neural} address discontinuities through Poisson jumps, while Stable SDEs~\cite{oh2024stable} introduce variants like Langevin-type, linear-noise, and geometric models to enhance the analysis of irregular time series. Collectively, these developments position Neural SDEs as a versatile and robust framework for modeling complex stochastic systems in continuous time. A detailed overview is presented in \cref{tab:De_dN} for Neural SDEs and their variants.

The evolution of Neural SDEs embodies a critical trade-off between uncertainty quantification and optimization tractability. Theoretically, Neural SDEs succeed exactly where deterministic ODEs fail: their learnable diffusion terms break rigid topological constraints to naturally capture aleatoric uncertainty and robustly model highly irregular, noisy data. However, this stochastic expressivity introduces severe computational failures. The inherent non-differentiability of Brownian motion forces stochastic integrators to use vanishingly small steps, while computing gradients via the stochastic adjoint method frequently yields catastrophic variance, destabilizing optimization. Recent advancements succeed by mathematically circumventing these numerical bottlenecks. Techniques such as signature kernels, cubature methods, and deterministic approximations shift the objective from tracking high-variance individual paths to matching robust topological signatures or finite-dimensional distributions. Ultimately, deploying Neural SDEs requires strictly weighing the necessity of modeling intrinsic randomness against the steep costs of numerical integration and gradient stabilization.

\subsubsection{\textbf{SDE-Guided Generative Models}~\cite{ho2020denoising,songscore}}
SDE-guided generative models generate samples by first adding noise to data in a forward process, then reconstructing data from noise in a reverse denoising process. The forward process can be formulated as a SDE:
\begin{equation}\label{SDE2}
    dx = f(x,t)\,dt+g(x,t)\,d\mathcal{W}(t),
\end{equation}
where $f(x,t)=-\frac12\beta(t)x$ and $g(t)=\sqrt{\beta(t)}$ in the standard DDPMs setting. In DDPMs, the generative process is realized by solving the reverse SDE, which simulates the dynamic evolution from the noise distribution to the target data distribution. The evolution of the system's probability density is described by the FPE, which accurately characterizes the statistical properties of the underlying stochastic dynamics. The corresponding reverse SDE can be expressed as:
\begin{equation}\label{RSDE1}
\begin{aligned}
dx \!=\! \Big[ f(x,t)\!-\!g^2(t)\nabla_x \log p(x,t)\Big] dt\!+\! g(t) d\overline{\mathcal{W}}(t),
\end{aligned}
\end{equation}
where the score function $\nabla_x \log p(x,t)$ is typically approximated by a parameterized neural network $s_\theta(x,t)$ and trained using score matching methods. Therefore, the generative process of DDPMs essentially involves the discrete numerical solution of the aforementioned reverse SDE.

Guided by the DDPMs paradigm, the development of SDE-guided generative models can be systematically divided into three major directions: the design of forward SDEs, the design of reverse SDEs, and the development of solvers for reverse SDEs.

\textbf{a) Forward SDE:}
In diffusion models, the forward process describes the transformation of the data distribution into a prior noise distribution. This process is typically represented by a SDE, as shown in Equation \eqref{SDE2}. By flexibly selecting specific forms for the drift term \( f(x, t) \) and the diffusion term \( g(t) \), or by extending the diffusion term to a state-dependent function \( g(x, t) \) to include multiplicative noise, various diffusion model variants can be developed. Several representative works will be introduced in the following section.

\textbf{Mean-Reverting SDE (MRSDE)}~\cite{luo2023image} extends the diffusion framework to model processes with a natural tendency to revert to a long-term mean, making it suitable for mean-reverting scenarios. The process is described by:
\[dx(t)=\theta(t)(\mu-x)dt+g(t)d\mathcal{W}(t),\]
which captures mean-reverting trajectories and stabilizes long-term generative dynamics. Beyond classical additive noise SDEs, various forward processes inject Gaussian noise using distinct schedules and parameterizations. Notable examples include Noise Conditional Score Networks (NCSNs)~\cite{song2019generative}, Multi-Speed Diffusion (MSDiff)~\cite{batzolis2021conditional}, Bayesian Flow Networks (BFNs)~\cite{xue2024unifying}, and Ornstein-Uhlenbeck diffusion~\cite{cao2025generative}.

When the diffusion term $g(x, t)$ depends on $x$, the additive SDE framework extends to multiplicative noise SDEs, enabling the modeling of state-dependent stochasticity. This extension addresses more complex phenomena such as variable uncertainty or structural constraints in data~\cite{richemond2022categorical,du2023flexible}.

\textbf{Flexible Diffusion Processes (FP-Diffusion)}~\cite{du2023flexible} enhances model expressiveness by simultaneously learning the drift $f(x)$ and data-dependent diffusion $R(x)$ through a multiplicative noise SDE, surpassing traditional additive noise methods:
\begin{equation*}
    dx = f(x)\beta'(t)dt+\sqrt{2\beta'(t)R(x)}d\mathcal{W}(t),
\end{equation*}
which allows the model to adapt flexibly to heterogeneous data distributions and to capture more intricate generative dynamics by learning both the drift and diffusion structures.

Recent works have generalized the framework by considering non-Gaussian noise~\cite{yoon2023score,voleti2022score,huang2024blue,vandersanden2025edgepreserving,shariatian2024denoising}, which broadens the class of data distributions that can be modeled. Further extensions introduce physical constraints to the diffusion process, for example, by enforcing boundary conditions or reflecting barriers~\cite{yu2023high,yuan2023physdiff,lu2023speed,lou2023reflected,nie2023blessing}.

\textbf{b) Reverse SDE:}
While the reverse SDE \eqref{RSDE1} provides a principled trajectory, its strict dependence on the forward process restricts modeling flexibility. Consequently, recent studies mathematically redesign the reverse SDE by learning more expressive drift functions or jointly optimizing the drift and diffusion terms. \textbf{Unified Framework}~\cite{cao2023exploring} introduces a tunable parameter $\gamma \in [0, 1]$ into the reverse SDE, allowing smooth interpolation between deterministic ODEs and stochastic SDEs:
\[dx=[f(x,t)-\gamma\cdot\frac12 g^2(t) s_\theta(x,t)]dt+\sqrt{1-\gamma^2}g(t)d\overline{\mathcal{W}}(t),\]
which can flexibly balance between sample diversity (stochasticity) and sample fidelity (determinism), and unifies a wide range of generative modeling dynamics within a single formulation.

This perspective connects reverse-SDE formulations with ODE-based flow models. Flow matching~\cite{lipman2022flow} and rectified flow~\cite{liu2022rectified} rely on deterministic velocity fields to construct efficient transport dynamics, whereas reverse-SDE formulations retain diffusion terms to model stochasticity and improve distributional coverage. Hybrid ODE/SDE dynamics may thus be understood as a continuum between deterministic transport and stochastic diffusion, in which the drift or velocity term characterizes the transport geometry and the diffusion term modulates the stochastic component. Other recent advances, such as DiffFlow~\cite{zhang2021diffusion}, DEIS~\cite{zhang2022fast} and ER SDE~\cite{cui2025elucidating}, further explore hybrid drift formulations and independent tuning of drift and diffusion terms, respectively. These extensions provide additional flexibility for balancing sample quality, diversity, and robustness in high-variance regions.

\textbf{c) Solvers for reverse SDE:}
Beyond redesigning the reverse dynamics, sampling inherently relies on integrating learned ODEs or SDEs. Consequently, advanced numerical solvers can directly accelerate generation and enhance precision without altering the underlying model architecture. A variety of solvers have been proposed to approximate the solution of the reverse SDEs or its deterministic ODEs counterpart. For example, the reverse process in Probability Flow ODE (PF ODE)~\cite{song2020score} can be formulated as the following ODE:
\begin{equation}
dx = [f(x, t) - \frac12 g^2(t) s_\theta(x, t)]dt.
\end{equation}
Several representative solvers have been developed to improve sampling efficiency and quality in practice. Denoising Diffusion Implicit Models (DDIMs)~\cite{song2020denoising} can be interpreted as a deterministic numerical solver for the underlying probability flow ODE, accelerating sampling. Generalized denoising diffusion implicit models (gDDIM)~\cite{zhang2022gddim} interpolates between deterministic and stochastic sampling.  Diffusion Probabilistic Model Solver (DPM-Solver)~\cite{lu2022dpm,lu2025dpm}, Pseudo Numerical methods for Diffusion
models (PNDMs)~\cite{liu2022pseudo}, Approximate Mean-Direction Solver (AMED-Solver)~\cite{zhou2024fast}, Stochastic Runge-Kutta Methods (SRK)~\cite{wu2024stochastic} and Stochastic Adaptive Solver (SA-Solver)~\cite{xue2023sa} apply high-order integration methods to improve sampling efficiency and quality. These solvers enable fast and diverse sample generation with fewer steps. Beyond generation, integrating DDPMs' stochastic denoising into training acts as a effective regularizer, enhancing stability and generalization while bypassing iterative inference in downstream tasks like image inpainting \cite{liao2025denoising} and segmentation \cite{guo2025take}.

SDE-guided generative models succeed over deterministic continuous flows by utilizing stochastic noise as a smoothing operator to break topology-preserving constraints, guaranteeing robust coverage of complex data distributions. However, this stochastic success introduces a inference bottleneck: the reverse denoising SDE generates highly curved transport trajectories that cause massive truncation errors during discretization, forcing basic solvers into thousands of costly evaluation steps. To overcome these integration bottlenecks, research progresses along two complementary trajectories. First, advanced models like FP-Diffusion~\cite{du2023flexible} and the Unified Framework~\cite{cao2023exploring} redesign the governing dynamics to balance deterministic transport with stochastic exploration, thereby reducing reverse-path curvature while preserving the distributional coverage benefits of diffusion. Second, high-order numerical solvers including DPM-Solver~\cite{lu2022dpm} and SA-Solver~\cite{xue2023sa} are deployed to directly combat curvature-induced errors. Ultimately, this reveals a fundamental computational trade-off: optimizing diffusion models requires mitigating reverse-path geometric curvature to maximize sampling efficiency without sacrificing the vital stochastic smoothing benefits of the forward process.

\section{Applications and Performance}
This section reviews the experimental performance of a range of DE-driven neural networks on CIFAR-10 and CIFAR-100
benchmarks. While Neural ODEs and Neural SDEs particularly excel in modeling sequential and time-series data,
we refer the reader to ~\cite{oh2025comprehensive,niu2024applications} for detailed results in that domain. Here, our focus is on architectures and generative
models informed by deterministic DEs and SDEs. We structure our discussion as follows: We first review the experimental setup used in the surveyed CIFAR-based studies, and then conduct a qualitative analysis of deterministic DE-driven neural architectures evaluated primarily on CIFAR-10 and CIFAR-100, as summarized in \cref{tab:de-nn}, alongside ODE-guided and SDE-guided generative models evaluated mainly on CIFAR-10, as detailed in \cref{tab:diffusion-nn}. Section S3 of the online supplementary material complements this focused performance discussion by summarizing representative tasks, application domains, and benchmark datasets for the three model families in Tab. S1, Tab. S2, and Tab. S3. These appendix tables serve as applicability-oriented references rather than additional quantitative comparisons. Through a comprehensive analysis of empirical results across these categories, we clarify the advantages and drawbacks of different models, providing actionable insights for designing equation-inspired architectures with improved training stability and task performance.

\subsection{Experiment setup} For CIFAR experiments, a standard data augmentation ($4$-pixel padding followed by $32\times32$ random cropping or horizontal flipping) is uniformly applied, and pixel values are rescaled to $[0, 1]$. The experimental configurations in the surveyed literature exhibit significant diversity in optimization trajectories. 

Standard classification tasks \cite{zhang2019multiple, zhu2023convolutional, lu2018beyond} typically utilize a $300$-epoch regimen with an initial learning rate (LR) of $0.1$ and a mini-batch size ranging from $32$ to $128$. In contrast, alternative strategies have been explored to enhance convergence or stability: a staged decay over 80k steps is employed in \cite{chang2018reversible}, whereas the Adam optimizer with a fixed learning rate of $0.001$ is used for $1000$ epochs in \cite{perugachi2021invertible}. Furthermore, piecewise constant schedules and momentum-based constant LR strategies are adopted in \cite{ruthotto2020deep} and \cite{sahakyan2023enhancing}, spanning $100$ to $200$ epochs to accommodate specific architectural constraints.

Generative frameworks \cite{tang2024contractive, lou2023reflected} follow the specialized training protocols established by \cite{song2019generative}. These models are typically optimized for $1.3$M iterations, with specific SDE variants requiring only $50000$, using the Adam optimizer with a mini-batch size of $128$ and a constant learning rate of $0.001$.  To capture multi-scale score functions, a geometric sequence of 10 noise scales $\{\sigma_i\}_{i=1}^{10}$ is configured, ranging from $1$ to $0.01$. During the sampling phase via annealed Langevin dynamics, $T=100$ steps and a step size of $\epsilon=2\times10^{-5}$ are adopted. Recent advancements in consistency and flow-matching models further optimize these trajectories. Liu et al. \cite{liu2025learning} and Ma et al. \cite{ma2025learning} employ the RAdam optimizer with mini-batch sizes ranging from $128$ to $512$ over $100$k iterations. Zhou et al. \cite{zhou2024fast} train a lightweight predictor on $10$k images using a polynomial time schedule, while Zhang et al. \cite{zhang2025hierarchical} adopt an effective batch size of $256$ for $1000$ epochs (approximately $391$k iterations) with a polynomial decay schedule and a $45$k-step warmup phase.

\subsection{Deterministic DE-Driven Network Architectures}
 We provide the experimental results of networks guided by first-order ODEs, higher-order ODEs, as well as systems of equations and PDEs in \cref{tab:de-nn}. Consistent with convergence order theory in numerical analysis, networks employing higher-order or more sophisticated schemes generally achieve superior performance.

The experimental results highlight a clear progression, where innovations in DEs modeling translate into advances in both network architectures and empirical performance. ResNets exemplify the explicit Euler discretization of first-order ODEs, where the skip connection directly mirrors the explicit integration step, achieving a strong baseline accuracy of $94.07\%$ on CIFAR-10 and $74.84\%$ on CIFAR-100. As variants of ResNet, RiR and ResNeXt further improve accuracy by exploring different formulations. With further advances in mathematical formulation, FractalNet employs a multi-path architecture inspired by Runge-Kutta integration, while RKCNN adopts the fourth-order Runge-Kutta method. These higher-order discretizations enable more accurate modeling of feature dynamics, resulting in notable improvements, such as RKCNN’s $95.52\%$ on CIFAR-10 and $77.31\%$ on CIFAR-100. Furthermore, LM-ResNets and LM-ResNeXt employ a linear multi-step strategy to significantly enhance the accuracy of first-order networks under the same configurations. 

Extending beyond ODEs, PDEs introduce spatiotemporal priors whose benefits scale with model capacity, as detailed in \cref{tab:de-nn}. Lightweight architectures such as Hamiltonian CNNs~\cite{ruthotto2020deep} utilize sub-million parameter budgets to prioritize stability and interpretability over absolute accuracy. Conversely, at standard capacities, PDE networks yield explicit empirical gains. Specifically, $11.17$M-parameter Parabolic and Hyperbolic PDE models \cite{sahakyan2023enhancing} achieve up to $95.70\%$ on CIFAR-10, outperforming the $94.47\%$ ResNet baseline. Thus, PDE modeling provides mathematical stability for resource-constrained scenarios and superior accuracy for scaled networks.

Overall, DE-driven architectures elevate network design from heuristic discrete mappings to theoretically grounded continuous dynamical systems. This paradigm provides a principled architectural toolkit: first-order ODEs yield computational efficiency, higher-order dynamics capture complex non-linearities, and PDEs enforce rigorous mathematical stability and spatial interactions. Collectively, these continuous formulations significantly enhance network expressiveness and accuracy, offering profound insights for modern model design.

{
\renewcommand{\arraystretch}{0.7}  
\setlength{\tabcolsep}{1.0pt}        
\scriptsize 
\begin{longtable}{p{2.5cm} p{2.0cm} p{1.5cm} p{1.5cm} p{1.5cm} p{1.5cm}}
\caption{Performance comparison of representative deterministic DE-driven neural network architectures on CIFAR-10 and CIFAR-100. Each methodological sub-block (First-order, Higher-order, DEs, or PDE-guided) indicates its source references sequentially, following the order in which the results appear. Within the same sub-block, results from different references are separated by vertical parallel lines.}
\label{tab:de-nn} \\
\toprule
\textbf{Model} & \textbf{Publication} & \textbf{Code} & \textbf{Params} & \textbf{Dataset} & {\textbf{Test acc. (\%)}} \\
\midrule
\endfirsthead
\toprule
\textbf{Model} & \textbf{Publication} & \textbf{Code} & \textbf{Params} & \textbf{Dataset} & {\textbf{Test acc. (\%)}} \\
\midrule
\endhead
\bottomrule
\endfoot
\addlinespace[0.3em]
\multicolumn{6}{c}{\textbf{(a) First-order Guided \cite{zhang2019multiple} \cite{zhu2023convolutional}}} \\
\addlinespace[0.3em]
\midrule
\multirow{2}{*}{ResNets \cite{he2016deep}} 
& \multirow{2}{*}{CVPR 2016} 
& \multirow{2}{*}{\href{https://github.com/KaimingHe/deep-residual-networks}{\textcolor{blue}{[code]}}} 
& 2.5M & CIFAR-10 & 94.07 \\
& & & 2.5M & CIFAR-100 & 74.84 \\
\multirow{2}{*}{RiR \cite{targ2016resnet}} 
& \multirow{2}{*}{ICLR 2016} 
& \multirow{2}{*}{\href{https://github.com/nutszebra/resnet_in_resnet}{\textcolor{blue}{[code]}}} 
& 10.3M & CIFAR-10 & 94.99 \\
& & & 10.3M & CIFAR-100 & 77.10 \\
\multirow{2}{*}{ResNeXt \cite{xie2017aggregated}} 
& \multirow{2}{*}{CVPR 2017} 
& \multirow{2}{*}{\href{https://github.com/facebookresearch/ResNeXt}{\textcolor{blue}{[code]}}} 
& 68.1M & CIFAR-10 & 96.42 \\
& & & 68.1M & CIFAR-100 & 82.69 \\
\multirow{2}{*}{FractalNet \cite{cho2024operator}} 
& \multirow{2}{*}{ICLR 2017} 
& \multirow{2}{*}{\href{https://github.com/gustavla/fractalnet}{\textcolor{blue}{[code]}}} 
& 30M & CIFAR-10 & 95.41 \\
& & & 30M & CIFAR-100 & 77.15 \\
\midrule
\midrule
\multirow{2}{*}{ResNets} 
& \multirow{2}{*}{CVPR 2016} 
& \multirow{2}{*}{\href{https://github.com/KaimingHe/deep-residual-networks}{\textcolor{blue}{[code]}}} 
& 0.14M & CIFAR-10 & 91.98 \\
& & & 0.14M & CIFAR-100 & 68.97 \\
\multirow{2}{*}{RKCNN \cite{zhu2023convolutional}} 
& \multirow{2}{*}{{\makecell{Neural Comput. \\
Appl. 2023}} }
& \multirow{2}{*}{-} 
& 1.14M & CIFAR-10 & 95.52 \\
& & & 1.14M & CIFAR-100 & 77.31 \\
\midrule
\addlinespace[0.5em]
\multicolumn{6}{c}{\textbf{(b) High-order Guided \cite{lu2018beyond}}} \\
\addlinespace[0.3em]
\midrule
\multirow{2}{*}{ResNets} 
& \multirow{2}{*}{CVPR 2016} 
& \multirow{2}{*}{\href{https://github.com/KaimingHe/deep-residual-networks}{\textcolor{blue}{[code]}}} 
& 1.14M & CIFAR-10 & 93.63 \\
& & & 1.7M & CIFAR-100 & 72.24 \\
\multirow{2}{*}{ResNeXt} 
& \multirow{2}{*}{CVPR 2017} 
& \multirow{2}{*}{\href{https://github.com/facebookresearch/ResNeXt}{\textcolor{blue}{[code]}}}&- 
& CIFAR-10 & - \\
& & & 34.4M & CIFAR-100 & 82.23 \\
\multirow{2}{*}{FractalNet} 
& \multirow{2}{*}{ICLR 2017} 
& \multirow{2}{*}{\href{https://github.com/gustavla/fractalnet}{\textcolor{blue}{[code]}}} 
& - & CIFAR-10 & - \\
& & & 22.9M & CIFAR-100 & 77.51 \\
\multirow{2}{*}{DenseNet \cite{huang2017densely}} 
& \multirow{4}{*}{CVPR 2017} 
& \multirow{4}{*}{\href{https://github.com/liuzhuang13/DenseNet}{\textcolor{blue}{[code]}}} 
& - & CIFAR-10 & - \\
& & & 27.2M & CIFAR-100 & 80.75 \\
\multirow{2}{*}{DenseNet-BC \cite{huang2017densely}} 
& & & - & CIFAR-10 & - \\
& & & 15.3M & CIFAR-100 & 82.40 \\
\multirow{2}{*}{LM-ResNets\cite{lu2018beyond}} 
& \multirow{4}{*}{ICML 2018} 
& \multirow{4}{*}{\href{https://github.com/2prime/LM-ResNet}{\textcolor{blue}{[code]}}} 
&1.14M  & CIFAR-10 & 93.84 \\
& & & 1.7M & CIFAR-100 & 74.13 \\
\multirow{2}{*}{LM-ResNeXt \cite{lu2018beyond}} 
&&
&-  & CIFAR-10 & - \\
& & & 35.1M & CIFAR-100 & 82.51 \\
\midrule
\addlinespace[0.5em]
\multicolumn{6}{c}{\textbf{(c) DEs-Guided  \cite{chang2018reversible} \cite{perugachi2021invertible}}} \\
\addlinespace[0.3em]
\midrule
\multirow{2}{*}{ResNets} 
& \multirow{2}{*}{CVPR 2016} 
& \multirow{2}{*}{\href{https://github.com/KaimingHe/deep-residual-networks}{\textcolor{blue}{[code]}}}  & 0.46M & CIFAR-10  & 92.86 \\
& & & 0.47M & CIFAR-100 & 70.05 \\
\multirow{2}{*}{RevNets \cite{gomez2017reversible}} 
& \multirow{2}{*}{NIPs 2017} 
& \multirow{2}{*}{\href{ https://github.com/renmengye/revnet-public}{\textcolor{blue}{[code]}}} 
& 0.46M & CIFAR-10 & 92.76 \\
& & &0.48M  & CIFAR-100 & 71.04 \\
\multirow{2}{*}{HamiltonianNet \cite{chang2018reversible}} 
& \multirow{2}{*}{AAAI 2018} 
& \multirow{2}{*}{-}
& 0.43M & CIFAR-10 & 92.76 \\
& & &0.44M  & CIFAR-100 & 69.78 \\
\midrule
\midrule
\multirow{2}{*}{i-Resnets \cite{behrmann2019invertible}} 
& \multirow{2}{*}{ICML 2019} 
& \multirow{2}{*}{\href{https://github.com/jhjacobsen/invertible-resnet}{\textcolor{blue}{[code]}}} 
& 8.7M & CIFAR-10 & 96.55 \\
& & & - & CIFAR-100 & - \\
\multirow{2}{*}{i-DenseNet \cite{perugachi2021invertible}} 
& \multirow{2}{*}{NIPs 2021} 
& \multirow{2}{*}{\href{https://github.com/yperugachidiaz/invertible_densenets}{\textcolor{blue}{[code]}}} 
& 8.7M & CIFAR-10 & 96.63 \\
& & & - & CIFAR-100 & - \\
\midrule
\addlinespace[0.5em]
\multicolumn{6}{c}{\textbf{(d) PDE-Guided  \cite{ruthotto2020deep}\cite{sahakyan2023enhancing}}} \\
\addlinespace[0.3em]
\midrule
\multirow{2}{*}{\tiny{Parabolic CNNs} \cite{ruthotto2020deep}} 
& \multirow{6}{*}{JMIV 2020} 
& \multirow{6}{*}{\href{https://github.com/EmoryMLIP/DynamicBlocks}{\textcolor{blue}{[code]}}} 
&0.26M  & CIFAR-10  & 88.5 \\
& & &0.65M  & CIFAR-100  & 64.8 \\
\multirow{2}{*}{\tiny{Hamiltonian CNNs} \cite{ruthotto2020deep}} 
&&
&0.51M  & CIFAR-10  & 89.3 \\
& & &0.36M  & CIFAR-100  & 64.9 \\
\multirow{2}{*}{\tiny{Second-order CNNs }\cite{ruthotto2020deep}} 
&&
&0.51M  & CIFAR-10  & 89.2 \\
& & &0.65M  & CIFAR-100  & 65.4 \\
\midrule
\midrule
\multirow{2}{*}{ResNets} 
& \multirow{2}{*}{CVPR 2016} 
& \multirow{2}{*}{\href{https://github.com/KaimingHe/deep-residual-networks}{\textcolor{blue}{[code]}}} 
& 11.17M & CIFAR-10  & 94.47 \\
& & & - & CIFAR-100 & -  \\
\multirow{2}{*}{Parabolic PDE \cite{sahakyan2023enhancing}} 
& \multirow{4}{*}{\makecell{Program. Comput.\\ Softw. 2023}} 
&\multirow{4}{*}{-}
& 11.17M & CIFAR-10  & 95.42 \\
& & & - & CIFAR-100 & -   \\
\multirow{2}{*}{Hyperbolic PDE \cite{sahakyan2023enhancing}} 
&&
& 11.17M & CIFAR-10  & 95.70  \\
& & & - & CIFAR-100& -  \\
\end{longtable}}

\subsection{DE-guided generative models}
\cref{tab:diffusion-nn} presents a comprehensive taxonomy of representative generative models, systematically evaluating them based on their governing dynamical frameworks (i.e., ODE- versus SDE-guided paradigms). Over time, advancements in mathematical modeling and solver design have led to consistent improvements in both generative quality and sampling efficiency on CIFAR-10, as reflected by the trends in Fr\'{e}chet Inception Distance (FID) and Number of Function Evaluations (NFE) metrics.

Benchmark results on CIFAR-10 illustrate a shift in generative modeling from stochastic diffusion to highly efficient deterministic trajectories. Early models like DDPMs require substantial computational budgets exceeding $1000$ steps and experience severe fidelity degradation under sparse sampling, with FID escalating from $3.17$ to $137.7$ at $20$ steps. This failure primarily stems from the large truncation errors caused by highly curved stochastic paths. Specialized continuous-time formulations and numerical solvers help address this limitation. Score-based SDEs stabilize generation quality, while the EDM framework utilizes refined preconditioning to achieve a state-of-the-art FID of $1.96$ in only $35$ steps. Specialized ODE solvers, including DPM-Solver and AMED-Solver, further compress the required budget to under 10 steps while maintaining robust fidelity. Ultimately, methods such as Rectified Flow redefine efficiency limits by straightening transport paths, enabling near-linear velocity fields and high-quality one-step generation.

Despite these algorithmic accelerations, achieving ultra-fast inference introduces a strict trade-off among sample fidelity, inference latency, and training overhead. The transition to single-iteration models represents a redistribution rather than an elimination of computational burden. While classical SDEs maximize sample quality at the cost of high latency, advanced ODE solvers provide a highly practical balance by accelerating inference without requiring supplementary optimization. Conversely, single-step generation relies on computationally expensive offline distillation. Therefore, practical deployment dictates model selection: ODE solvers remain optimal for environments with constrained training resources that prioritize generation quality, whereas distillation approaches are suitable for latency-critical applications capable of absorbing large upfront training costs.

{\renewcommand{\arraystretch}{1.0}  
\setlength{\tabcolsep}{0.9pt}        
\scriptsize 
\begin{longtable}{p{3.7cm} p{2.0cm} p{1.5cm} p{1.5cm} p{1.5cm} p{1.5cm}}
\caption{Performance comparison of representative ODE-guided and SDE-guided generative models on CIFAR-10. Each methodological sub-block (Forward SDE, or Reverse SDE) indicates its source references sequentially, following the order in which the results appear. Within the same sub-block, results from different references are separated by vertical parallel lines.}
\label{tab:diffusion-nn}\\
\toprule
\textbf{Model} & \textbf{Publication} & \textbf{Code}  & \textbf{Dataset} &{\textbf{FID}{\textbf{$\downarrow$}}} &{\textbf{NFE}{\textbf{$\downarrow$}}}
\\
\midrule
\endfirsthead
\midrule
\endhead
\bottomrule
\endfoot
\addlinespace[0.3em]
\multicolumn{6}{c}{\textbf{(a) SDE-guided generative models: Forward SDE~\cite{tang2024contractive}\cite{lou2023reflected}}}\\
\addlinespace[0.3em]
\midrule
\multirow{1}{*}{NCSN \cite{song2019generative}} & \multirow{1}{*}{NIPs 2019} & \multirow{1}{*}{\href{https://github.com/ermongroup/ncsn}{\textcolor{blue}{[code]}}}   & CIFAR-10 &25.32 &1000\\
\multirow{1}{*}{NCSNv2 \cite{song2020improved}} & \multirow{1}{*}{NIPs 2020} & \multirow{1}{*}{\href{https://github.com/ermongroup/ncsnv2}{\textcolor{blue}{[code]}}}  & CIFAR-10 &10.87 &1160\\
\multirow{1}{*}{DDPMs \cite{ho2020denoising}} & \multirow{1}{*}{NIPs 2020} & \multirow{1}{*}{\href{https://github.com/hojonathanho/diffusion}{\textcolor{blue}{[code]}}}  & CIFAR-10 &3.17 &1000\\
\multirow{1}{*}{Score-based SDE-VP \cite{songscore}} & \multirow{1}{*}{ICLR 2021} & \multirow{1}{*}{\href{https://github.com/yang-song/score_sde_pytorch}{\textcolor{blue}{[code]}}}  & CIFAR-10 &2.55 &1000\\
\multirow{1}{*}{Score-based SDE-subVP \cite{songscore}} & \multirow{1}{*}{ICLR 2021} & \multirow{1}{*}{\href{https://github.com/yang-song/score_sde_pytorch}{\textcolor{blue}{[code]}}}  & CIFAR-10 &2.61 &1000\\
\multirow{1}{*}{Score-based SDE-VE \cite{songscore}} & \multirow{1}{*}{ICLR 2021} & \multirow{1}{*}{\href{https://github.com/yang-song/score_sde_pytorch}{\textcolor{blue}{[code]}}}  & CIFAR-10 &2.50 &1000\\
\multirow{1}{*}{EDM-VP \cite{Karras2022edm}} & \multirow{1}{*}{NIPs 2022} & \multirow{1}{*}{\href{https://github.com/NVlabs/edm}{\textcolor{blue}{[code]}}}   & CIFAR-10 &1.96 &35\\
\multirow{1}{*}{EDM-VE \cite{Karras2022edm}} & \multirow{1}{*}{NIPs 2022} & \multirow{1}{*}{\href{https://github.com/NVlabs/edm}{\textcolor{blue}{[code]}}}   & CIFAR-10 &1.97 &35\\
\midrule
\midrule
\multirow{1}{*}{DDPMs \cite{ho2020denoising}} & \multirow{1}{*}{NIPs 2020} & \multirow{1}{*}{\href{https://github.com/hojonathanho/diffusion}{\textcolor{blue}{[code]}}} & CIFAR-10 &10.61 &100\\
\multirow{1}{*}{DDPMs \cite{ho2020denoising}} & \multirow{1}{*}{NIPs 2020} & \multirow{1}{*}{\href{https://github.com/hojonathanho/diffusion}{\textcolor{blue}{[code]}}} & CIFAR-10 &35.29 &50\\
\multirow{1}{*}{DDPMs \cite{ho2020denoising}} & \multirow{1}{*}{NIPs 2020} & \multirow{1}{*}{\href{https://github.com/hojonathanho/diffusion}{\textcolor{blue}{[code]}}} & CIFAR-10 &137.7 &20\\
\multirow{1}{*}{FastDPM \cite{kong2021fast}} & \multirow{1}{*}{ICML 2021} & \multirow{1}{*}{\href{https://github.com/zhifengkong/FastDPM_pytorch}{\textcolor{blue}{[code]}}}  & CIFAR-10 &3.01 &100\\
\multirow{1}{*}{FastDPM \cite{kong2021fast}} & \multirow{1}{*}{ICML 2021} & \multirow{1}{*}{\href{https://github.com/zhifengkong/FastDPM_pytorch}{\textcolor{blue}{[code]}}}  & CIFAR-10 &3.41 &50\\
\multirow{1}{*}{FastDPM \cite{kong2021fast}} & \multirow{1}{*}{ICML 2021} & \multirow{1}{*}{\href{https://github.com/zhifengkong/FastDPM_pytorch}{\textcolor{blue}{[code]}}}  & CIFAR-10 &5.22 &20\\
\midrule
\midrule
\multirow{1}{*}{Score-based SDE \cite{songscore}} & \multirow{1}{*}{ICLR 2021} & \multirow{1}{*}{\href{https://github.com/yang-song/score_sde_pytorch}{\textcolor{blue}{[code]}}}   & CIFAR-10 &2.41 &1000\\
\multirow{1}{*}{EDM \cite{Karras2022edm}} & \multirow{1}{*}{NIPs 2022} & \multirow{1}{*}{\href{https://github.com/NVlabs/edm}{\textcolor{blue}{[code]}}}   & CIFAR-10 &1.97 &35\\
\multirow{1}{*}{Reflected SDE \cite{lou2023reflected}} & \multirow{1}{*}{ICML 2023} & \multirow{1}{*}{\href{https://github.com/louaaron/Reflected-Diffusion}{\textcolor{blue}{[code]}}}   & CIFAR-10 &2.72 &1000\\
\midrule
\midrule
\addlinespace[0.3em]
\multicolumn{6}{c}{\textbf{  SDE-guided generative models: Reverse SDE \cite{liu2025learning} \cite{zhou2024fast}}}\\
\addlinespace[0.3em]
\midrule
\multirow{1}{*}{CFM ~\cite{lipman2022flow}} & \multirow{1}{*}{ICLR 2023} & \multirow{1}{*}{\href{https://github.com/atong01/conditional-flow-matching}{\textcolor{blue}{[code]}}}   & CIFAR-10 &6.35 &142\\
\multirow{1}{*}{DEIS~\cite{zhang2022fast}} 
& \multirow{1}{*}{ICLR 2023} 
& \multirow{1}{*}{\href{https://github.com/qsh-zh/deis}{\textcolor{blue}{[code]}}} 
 & CIFAR-10 &2.55 &50\\
 \multirow{1}{*}{Rectified Flow ~\cite{liu2022rectified,liu2023flow}} & \multirow{1}{*}{ICLR 2023} & \multirow{1}{*}{\href{https://github.com/gnobitab/RectifiedFlow}{\textcolor{blue}{[code]}}}  & CIFAR-10 &2.58 &127\\
\multirow{1}{*}{DPM-Solver-2 \cite{lu2022dpm}} & \multirow{1}{*}{NIPs 2022} & \multirow{1}{*}{\href{https://github.com/LuChengTHU/dpm-solver}{\textcolor{blue}{[code]}}}   & CIFAR-10 &4.7 &10\\
\multirow{1}{*}{DPM-Solver-3 \cite{zheng2023dpm}}  & \multirow{1}{*}{NIPs 2023} & \multirow{1}{*}{\href{https://github.com/thu-ml/DPM-Solver-v3}{\textcolor{blue}{[code]}}}   & CIFAR-10 &2.51 &10\\
\multirow{1}{*}{DPM-Solver++ \cite{lu2025dpm}} & \multirow{1}{*}{MIR 2025} &\multirow{1}{*}{-}  & CIFAR-10 &2.91 &10\\
\midrule
\midrule
\multirow{1}{*}{DDIM \cite{song2020denoising}} & \multirow{1}{*}{ICML 2021} & \multirow{1}{*}{\href{https://github.com/ermongroup/ddim}{\textcolor{blue}{[code]}}}   & CIFAR-10 &18.43 &9\\
\multirow{1}{*}{DPM-Solver-2 \cite{lu2022dpm}} & \multirow{1}{*}{NIPs 2022} & \multirow{1}{*}{\href{https://github.com/LuChengTHU/dpm-solver}{\textcolor{blue}{[code]}}}   & CIFAR-10 &4.98 &9\\
\multirow{1}{*}{DPM-Solver++(3M) \cite{lu2025dpm}} & \multirow{1}{*}{MIR 2025}  & \multirow{1}{*}{-} & CIFAR-10 &3.42 &9\\
\multirow{1}{*}{iPNDM \cite{zhang2022fast}} 
& \multirow{1}{*}{ICLR 2023} 
& \multirow{1}{*}{\href{https://github.com/luping-liu/PNDM}{\textcolor{blue}{[code]}}} 
 & CIFAR-10 &3.17 &9\\
\multirow{1}{*}{AMED-Solver~\cite{zhou2024fast}} &\multirow{1}{*}{CVPR 2024} & \multirow{1}{*}{\href{https://github.com/IftikharIfti/AMED-solver-with-dpmpp}{\textcolor{blue}{[code]}}}  & CIFAR-10 &3.67 &9\\
\multirow{1}{*}{AMED-Plugin~\cite{zhou2024fast}} & \multirow{1}{*}{CVPR 2024} & \multirow{1}{*}{\href{https://github.com/IftikharIfti/AMED-solver-with-dpmpp}{\textcolor{blue}{[code]}}}   & CIFAR-10 &2.63 &9\\
\midrule
\addlinespace[0.3em]
\multicolumn{6}{c}{\textbf{(b) ODE-guided generative models \cite{zhang2025hierarchical}} \cite{ma2025learning}}\\
\addlinespace[0.3em]
\midrule
\multirow{1}{*}{Rectified Flow ~\cite{liu2022rectified,liu2023flow}} & \multirow{1}{*}{ICLR 2023} & \multirow{1}{*}{\href{https://github.com/gnobitab/RectifiedFlow}{\textcolor{blue}{[code]}}}  & CIFAR-10 &4.56 &100\\
\multirow{1}{*}{OT-CFM ~\cite{tong2024improving}} & \multirow{1}{*}{TMLR 2023} & \multirow{1}{*}{\href{https://github.com/atong01/conditional-flow-matching}{\textcolor{blue}{[code]}}}  & CIFAR-10 &4.95&100\\
\multirow{1}{*}{Hierarchical Rectified Flow~\cite{zhang2025towards}} & \multirow{1}{*}{ICLR 2025} & \multirow{1}{*}{\href{https://github.com/riccizz/HRF}{\textcolor{blue}{[code]}}}   & CIFAR-10 &4.33 &100\\
\multirow{1}{*}{HRF2~\cite{zhang2025hierarchical}} & \multirow{1}{*}{ICLR 2025} & \multirow{1}{*}{\href{https://riccizz.github.io/HRF_coupling}{\textcolor{blue}{[code]}}}   & CIFAR-10 &4.33 &100\\
\multirow{1}{*}{HRF2-D~\cite{zhang2025hierarchical}} & \multirow{1}{*}{ICLR 2025} & \multirow{1}{*}{\href{https://riccizz.github.io/HRF_coupling}{\textcolor{blue}{[code]}}}   & CIFAR-10 &4.30 &100\\
\midrule\midrule

\multirow{1}{*}{CFM ~\cite{lipman2022flow}} & \multirow{1}{*}{ICLR 2023} & \multirow{1}{*}{\href{https://github.com/atong01/conditional-flow-matching}{\textcolor{blue}{[code]}}}   & CIFAR-10 &166.65 &2\\

\multirow{1}{*}{1-Rectified Flow ~\cite{liu2022rectified,liu2023flow}} & \multirow{1}{*}{ICLR 2023} & \multirow{1}{*}{\href{https://github.com/gnobitab/RectifiedFlow}{\textcolor{blue}{[code]}}}  & CIFAR-10 &6.18 &2\\

\multirow{1}{*}{1-Rectified Flow ~\cite{liu2022rectified,liu2023flow}} & \multirow{1}{*}{ICLR 2023} & \multirow{1}{*}{\href{https://github.com/gnobitab/RectifiedFlow}{\textcolor{blue}{[code]}}}  & CIFAR-10 &378 &1\\

\multirow{1}{*}{3-Rectified Flow ~\cite{liu2022rectified,liu2023flow}} & \multirow{1}{*}{ICLR 2023} & \multirow{1}{*}{\href{https://github.com/gnobitab/RectifiedFlow}{\textcolor{blue}{[code]}}}  & CIFAR-10 &5.21 &2\\

\multirow{1}{*}{3-Rectified Flow ~\cite{liu2022rectified,liu2023flow}} & \multirow{1}{*}{ICLR 2023} & \multirow{1}{*}{\href{https://github.com/gnobitab/RectifiedFlow}{\textcolor{blue}{[code]}}}  & CIFAR-10 &8.15 &1\\
\multirow{1}{*}{VFM ~\cite{guo2025variational}} & \multirow{1}{*}{ICML 2025} & \multirow{1}{*}{-}   & CIFAR-10 &97.83 &2\\
\multirow{1}{*}{MeanFlow~\cite{geng2026mean}} & \multirow{1}{*}{NIPs 2025} & \multirow{1}{*}{\href{https://github.com/haidog-yaqub/MeanFlow}{\textcolor{blue}{[code]}}}  & CIFAR-10 &2.23  &2\\
\multirow{1}{*}{MeanFlow~\cite{geng2026mean}} & \multirow{1}{*}{NIPs 2025} & \multirow{1}{*}{\href{https://github.com/haidog-yaqub/MeanFlow}{\textcolor{blue}{[code]}}}  & CIFAR-10 &2.92  &1\\
\multirow{1}{*}{S-VFM~\cite{ma2025learning}} & \multirow{1}{*}{-} & \multirow{1}{*}{-}   & CIFAR-10 &2.16 &2\\
\multirow{1}{*}{S-VFM~\cite{ma2025learning}} & \multirow{1}{*}{-} & \multirow{1}{*}{-}   & CIFAR-10 &2.81 &1\\
\end{longtable}
}

\section{Discussions and Conclusion}
\subsection{Discussions}
The integration of DEs with deep learning has opened up several promising avenues for advancing the field, but significant opportunities and challenges remain \cite{hosain2024explainable,trigka2025comprehensive}. Building upon the empirical synthesis and fundamental limitations identified in our review, we argue that moving beyond empirical trial-and-error requires a deeper mathematical synthesis. As the relationship between DEs and deep learning evolves, future research should focus on the following key directions to address the limitations of current models.

\textbf{Overcoming Markovian Limitations.} Standard ODEs and CDEs often fail to capture long-range dependencies due to inherent Markovian assumptions. To mathematically embed historical memory and bypass these failures, future architectures should integrate higher-order~\cite{butcher2016numerical}, delay~\cite{erneux2009applied}, and fractional-order equations~\cite{chen2009fractional,kilbas2001differential}, which offer theoretically grounded solutions to memory decay.

\textbf{Breaking Gaussian Assumptions.} While SDEs successfully capture aleatoric uncertainty, they fail under extreme discontinuous shifts due to Gaussian assumptions. Exploring L\'evy processes~\cite{peszat2007stochastic} naturally accommodates heavy-tailed distributions and sudden jumps, providing rigorous regularization for highly volatile dynamics where standard continuous SDEs fail.

\textbf{Synthesizing ODE and SDE Paradigms.} Generative modeling highlights a trade-off between stochastic coverage and deterministic efficiency. SDE-based diffusion models improve distributional coverage and sample diversity through stochastic perturbations, but typically require iterative reverse-time integration. In contrast, ODE-based approaches, such as flow matching and rectified flow, learn deterministic velocity fields for efficient transport, but provide less explicit stochastic exploration. Recent hybrid formulations address this trade-off at the level of governing dynamics. For example, the Unified Framework~\cite{cao2023exploring} interpolates between probability-flow ODEs and reverse SDEs, while DEIS~\cite{zhang2022fast} and ER SDE~\cite{cui2025elucidating} adjust drift and diffusion terms to balance fidelity, diversity, and computational cost. This view also links naturally to flow-matching-based models, where velocity fields characterize transport geometry and diffusion terms modulate stochasticity. Future work may further develop principled ODE/SDE hybrid dynamics for efficient and robust generative modeling.

\textbf{Aligning with Domain Constraints.} The practical failure of DE-driven models often stems from the misalignment neural architectures with domain requirements~\cite{hua2025comprehensive,croitoru2023diffusion}. Because solving physical PDEs demands exact conservation laws~\cite{sewell2005numerical} while computer vision prioritizes robustness~\cite{szeliski2022computer}, future research should aim to explicitly embed mathematical properties (e.g., Lyapunov stability) into network designs to ensure theoretically rigorous and safely deployable models.

To sum up, these emerging directions underscore the immense potential of DEs in driving innovation within deep learning, guiding the development of models that are both mathematically principled and highly effective across a wide range of applications.

\subsection{Conclusion}
This survey provides a systematic review of the intersection between deep learning and DEs. Subsequently,  we establish a unified dual taxonomy that organizes the field by its mathematical formulations and methodological roles. Our analysis demonstrates that model success depends on the synergy between neural architectures and numerical schemes. We clarify the fundamental trade-offs among interpretability, stability, and computational cost to provide principled guidelines for framework selection. Furthermore, we elucidate how SDEs bridge discrete training with continuous dynamics for advanced uncertainty quantification and generative modeling.

Future research should focus on overcoming Markovian limitations by integrating delay and fractional-order equations, while relaxing restrictive Gaussian assumptions through L\'evy processes to model more volatile dynamics. Another promising direction is to design hybrid dynamics and align neural architectures with domain-specific constraints, thereby supporting the development of mathematically grounded, reliable, and adaptable intelligent systems. This structured perspective provides researchers with a clearer foundation for future innovation in this evolving field.

\begin{acks}
This work was supported by the 
\grantsponsor{NSFC}{National Natural Science Foundation of China (NSFC)}{} under Grant No.~ T2541053.
\end{acks}

{
\bibliographystyle{ACM-Reference-Format}
\bibliography{ref}
}

\end{document}